\documentclass[journal]{IEEEtran}
\ifCLASSINFOpdf
\else
\fi

\usepackage{array}
\usepackage{booktabs}
\usepackage{multirow}
\usepackage{amsmath}
\usepackage{amssymb}
\usepackage{amsthm}
\usepackage{graphicx}
\usepackage{float}
\usepackage{eucal}
\usepackage{cite}
\usepackage{subfigure}
\usepackage{url}
\usepackage{algorithm}
\usepackage{algorithmicx}
\usepackage{algpseudocode}
\usepackage{bm}
\usepackage{mathptmx}
\DeclareMathAlphabet{\mathcal}{OMS}{cmsy}{m}{n}
\DeclareSymbolFont{largesymbols}{OMX}{cmex}{m}{n}

\newtheorem{theorem}{Theorem}
\newtheorem{definition}{Definition}

\begin{document}
%
\title{Joint Projection Learning and Tensor Decomposition Based Incomplete Multi-view Clustering}
%
%

\author{Wei~Lv,
        Chao~Zhang,
        Huaxiong~Li,
        Xiuyi Jia, and~Chunlin~Chen~
\thanks{ This work is partially supported by the National Natural Science Foundation of China under Grants 62176116, 62073160, 62276136, and the Natural Science Foundation of the Jiangsu Higher Education Institutions of China under Grant 20KJA520006.
\textit{(Corresponding author: Chao Zhang.)}}
\thanks{Wei Lv, Chao Zhang, Huaxiong Li, and Chunlin Chen are with the Department of Control Science and Intelligence Engineering, Nanjing University, Nanjing 210093, China.  (e-mail: weilv@smail.nju.edu.cn; chzhang@smail.nju.edu.cn; huaxiongli@nju.edu.cn; clchen@nju.edu.cn).}
\thanks{Huaxiong Li is also with the Nanjing Institute of Agricultural Mechanization, Ministry of Agriculture and Rural Affairs, Nanjing 210014, China.}
	\thanks{Xiuyi Jia is with the School of Computer Science and Engineering,
		Nanjing University of Science and Technology, Nanjing 210014, China.
		(e-mail: jiaxy@njust.edu.cn).}
}

%
%

\markboth{IEEE TRANSACTIONS ON NEURAL NETWORKS AND LEARNING SYSTEMS}%
{IEEE TRANSACTIONS ON NEURAL NETWORKS AND LEARNING SYSTEMS}
%



\maketitle

\begin{abstract}
Incomplete multi-view clustering (IMVC) has received increasing attention since it is often that some views of samples are incomplete in reality.
Most existing methods learn similarity subgraphs from original incomplete multi-view data and seek complete graphs by exploring the incomplete subgraphs of each view for spectral clustering. However, the  graphs constructed on the original high-dimensional data may be suboptimal due to feature redundancy and noise. 
Besides, previous methods generally ignored the graph noise caused by the inter-class and intra-class structure variation during the transformation of incomplete graphs and complete graphs.
To address these problems,
we propose a novel Joint Projection Learning and Tensor Decomposition Based method (JPLTD) for IMVC. 
Specifically, to alleviate the influence of redundant features and noise in high-dimensional data, JPLTD introduces an orthogonal projection matrix to project the high-dimensional features into a lower-dimensional space for compact feature learning.
Meanwhile, based on the lower-dimensional space, the similarity graphs corresponding to instances of different views are learned, and JPLTD stacks these graphs into a third-order low-rank tensor to explore the high-order correlations across different views. 
We further consider the graph noise of projected data caused by missing samples and use a tensor-decomposition based graph filter for robust clustering.
JPLTD decomposes the original tensor into an intrinsic tensor and a sparse tensor.  The intrinsic tensor models the true data similarities.
An effective optimization algorithm is adopted to solve the JPLTD model. 
Comprehensive experiments on several benchmark datasets demonstrate that JPLTD outperforms the state-of-the-art methods. The code of JPLTD is available at https://github.com/weilvNJU/JPLTD.
\end{abstract}


%
\IEEEpeerreviewmaketitle

\section{Introduction}
%
%
%
%
\IEEEPARstart{I}{n}
real applications, multi-view data are universal. For instance, we can extract various types of features to describe an object. Multi-view data from different sources can better represent the semantic characteristics of objects than single-view data~\cite{1,2,3}. The universal multi-view data has aroused  extensive research interest in multi-view learning, in which multi-view clustering (MVC) is an important task. MVC explores the latent information across views to divide data samples into separate clusters without label annotations. In recent years, numerous MVC approaches have been investigated and presented,
which can be roughly divided into co-training strategy~\cite{4,5}, non-negative matrix factorization (NMF)~\cite{6,7,8,9}, multiple kernel clustering (MKC)~\cite{10,fR2,12,13,14} and graph-based clustering~\cite{17,18,19, addWu}.
The co-training methods for multi-view clustering aim to iteratively obtain clustering results of each view, and provide help for the clustering process of other views~\cite{4,5}.
The NMF approaches apply the matrix factorization technology to MVC. As a representative,
Zhang \textit{et al.}~\cite{8} introduced matrix factorization technology into MVC and presented a latent multi-view subspace clustering method (LSMC). LSMC seeks a unified latent representation of different views and improves the subspace clustering. 
The MKC methods try to optimize the coefficients of a group of pre-defined kernels and achieve optimal information fusion.
In~\cite{14}, Liu \textit{et al.} presented a method based on MKC named optimal neighborhood kernel clustering (ONKC). ONKC expands the feasible set of optimal kernels and pays attention to the negotiation between clustering and kernel learning.
The graph-based clustering methods aim to seek a consensus graph to characterize the pairwise similarities of data points.
Nie \textit{et al.}~\cite{17}  introduced a novel implicit weight learning mechanism and applied it to the constrained Laplacian rank  method, which aims to find a consensus Laplacian rank-constrained graph to connect all views. In addition, many deep multi-view clustering approaches have also been proposed to improve the clustering performance~\cite{20,21,22,23}. Although the above methods have achieved good performance, it should be noted that most of them assume the samples among each view are complete. Unfortunately, in reality,  it is common that the multi-view data suffers from several missing views. For example, people are often reluctant to input their personal information such as age, height, and weight into a mobile application, which will result in sample incompleteness in corresponding views. In this case, traditional MVC methods can hardly handle these incomplete multi-view data, which makes researchers begin to pay attention to incomplete multi-view clustering (IMVC).

To handle the IMVC problem, a series of approaches have been presented~\cite{recommand1}, most of which can be categorized into four aspects: kernel-based learning~\cite{25,26,27,28}, subspace-based learning~\cite{35,36, 38,40}, deep learning~\cite{41, 43,44}, and graph-based learning~\cite{29,recommand5,30,31,33,34,45,42,recommand2,recommand3, fR1}. The kernel-based IMVC approaches aim to impute
missing entries of the kernel matrix for clustering. Shao \textit{et al.}~\cite{25} presented a method named  Collective Kernel Learning (CoKL) to  collectively impute the kernel matrix of different datasets by  utilizing the shared samples across all datasets. CoKL imputes the incomplete kernels first and then obtains the clustering results based on the learned kernels. To improve the clustering efficiency, Liu \textit{et al.}~\cite{28} integrated clustering and kernel imputation into a unified learning procedure. They proposed two algorithms to address the incomplete multiple kernel clustering problem. The first one tries to impute the incomplete base kernels and then utilizes the standard MKC algorithm for clustering. The second method considers the correlation among incomplete base kernels and encourages them to complete each other. 
The subspace-based methods attempt to seek a latent subspace representation among all views for clustering. 
In~\cite{36}, Wen \textit{et al.} presented a  unified embedding alignment framework (UEAF) for IMVC. Instead of utilizing suboptimal two-phase learning, UEAF simultaneously completes the missing samples and seeks the latent representation of all views. Besides, it utilizes a 
graph regularization term to ensure the consistency of the local structure of various views. However, UEAF lacks consideration of complementary information across different views.
To address this problem,  
Li \textit{et al.}~\cite{40} presented a high-order correlation preserved incomplete multi-view subspace clustering (HCP-IMSC) approach, which introduces a  tensor factorization regularization and constructs a hypergraph to investigate the high-order property of 
multi-view data. 
Recently, some researchers adopt deep neural networks to handle the IMVC problem. As a representative, Xu \textit{et al.}~\cite{44} presented an imputation-free and fusion-free deep IMVC (DIMVC) method, which aims to explore the linear separability of different views by mapping their latent features into a high-dimensional space.

Since the graph-based IMVC methods can reveal the complex distribution of data and correlations among data, they
have raised increasing attention. The graph-based IMVC approaches generally try to learn
similarity graphs across multiple views and utilize index
matrices to link the subgraphs with complete graphs.
As a representative, Wen \textit{et al.}~\cite{29} presented an incomplete multiview spectral clustering method with adaptive graph learning (IMSC-AGL), which adopts the low-rank representation technique to excavate the low-rank structure of subgraphs. IMSC-AGL directly uses the existing data to learn subgraphs and its complete graphs are obtained by using index matrices to stretch subgraphs. However, IMSC-AGL ignores the consistency and specificity embedded in multi-view data. To consider this information, Lv \textit{et al.}~\cite{45} proposed a view-consistency learning for incomplete multi-view clustering (VCL-IMVC) approach. VCL-IMVC divides samples of each view into view-unimportant-content and view-important-content, then the affinity matrices of different views are learned according to their importance. 
Similar to most graph-based methods, the above two methods will encounter significant computational and  time costs when facing large-scale datasets.
To deal with large-scale IMVC tasks more effectively, Wang \textit{et al.}~\cite{30} introduced the consensus bipartite graph framework to deal with the IMVC problem (IMVC-CBG), which jointly learns consensus anchors and bipartite graphs 
for IMVC.
It should be noted that all of the aforementioned graph-based methods lack consideration of the high-order relationships of various views.
To address this problem, some researchers stack the complete graphs into a tensor and introduce low-rank tensor learning in graph-based IMVC.
Li \textit{et al.}~\cite{34} presented a tensor-based multi-view block diagonal structure diffusion (TMBSD) approach, which first uses the original samples to learn similarity subgraphs, and then attempts to complete a low-rank tensor to seek a common representation for clustering. Similar to TMBSD, the tensor completion-based incomplete multi-view clustering (TC-IMVC)~\cite{31} method also separates the graph learning and graph recovery process, and tries to recover the predefined graph. This two-phase learning strategy causes their clustering performance sensitive to the quality of initial graphs and makes it difficult to ensure the final graph quality. 
To address this problem, 
Wen \textit{et al.}~\cite{33} proposed an approach called incomplete multi-view tensor spectral clustering with missing view inferring (IMVTSC-MVI), which incorporates the missing-view inferring and graph learning into a unified framework. IMVTSC-MVI aims to utilize the original feature data and restore the missing views for graph learning.
Different from IMVTSC-MVI,
Li \textit{et al.}~\cite{42} believed that it is difficult for IMVC methods to recover the missing features in the original data space, therefore, they directly used existing data for graph learning and proposed a graph structure refining method for IMVC (GSRIMC).

 \begin{figure}[!t]
	\centering
	\includegraphics[width=8.67cm,height=6.5cm]{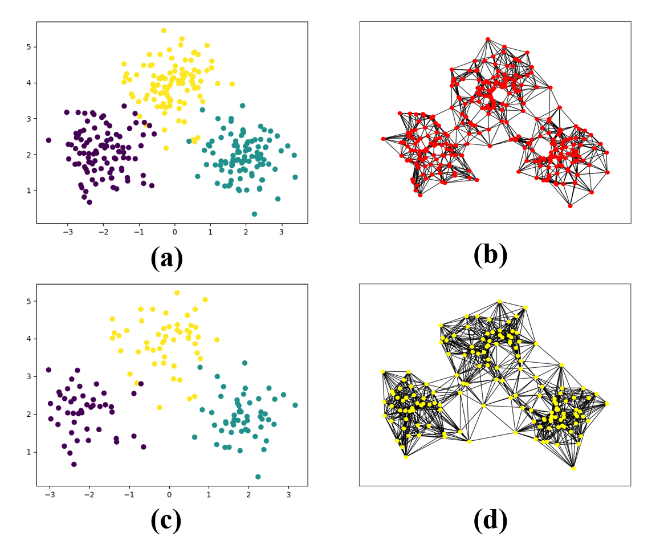}
	\caption{An example to explain the graph structure changes when dealing with complete data and incomplete data. (a) Complete data: 300 sample points from three classes; (b) The complete $k$NN graph ($k$=10) corresponding to (a); (c) Incomplete data: 150 sample points randomly selected from (a); (d) The incomplete graph corresponding to (c).}
	\label{fig1}
\end{figure}
In view of the existing graph-based approaches, it can be observed that: 
(1) Some previous IMVC methods attempt to  recover a complete graph from pre-defined graphs, which divide the graph construction and graph recovery into two individual steps, and thus the recovered graphs may be not globally optimal.
(2) Most methods usually constructed the initial graphs by computing the distances among original samples, which ignores the data redundancy and noise, leading to suboptimal graphs. 
(3) Most existing methods use index matrices to connect subgraphs and entire graphs, and let subgraphs guide the entire graph learning, ignoring the adverse effects of biased structure of subgraphs.
Fig.~\ref{fig1} exhibits an example to explain the biased structure of subgraphs. As shown in Fig.~\ref{fig1}(a), we randomly generate 300 sample points in three classes, (b) uses $k$=10 to show the $k$NN graph structure, (c) randomly removes half of the sample points in (a), and its corresponding $k$NN graph ($k$=10) structure is displayed in (d). We can observe that compared with (b), the edges among different classes in (d) are denser, indicating the inter-class correlations are closer, and making partitions for different classes is more difficult. Meanwhile,  due to the reduction of data points, the edges within the class are more clear, and the intra-class relations become sparser. 
In other words, both inter-class and intra-class correlations among corresponding data points have changed in the subgraphs. Directly using these subgraphs to guide the complete graphs learning may be suboptimal.
We regard this harmful information in learned graphs which will lead to suboptimal results as sparse noise. 

To address these issues, we propose a Joint projection learning and Tensor Decomposition (JPLTD) based method for IMVC. JPLTD integrates graph construction and tensor based graph recovery into a unified framework,  making the two components mutually boost. 
Specifically, 
to reduce the feature noise interference in the original high-dimensional data, JPLTD  performs joint projection learning and graph construction.
JPLTD projects the original data in each view into a lower-dimensional space for compact feature learning.
Meanwhile, JPLTD learns the similarity graph corresponding to each view based on the low-dimension space.  And the learned graphs are stacked into a third-order tensor for exploring the relationship across views. It should be noted that these learned graphs are corrupted and the formed tensor contains sparse noise in incomplete graphs.  To reduce the influence of the graph noise caused by missing samples and better characterize the similarities of existing data, we decompose the original tensor into a sparse tensor and an intrinsic tensor and apply a sparse constraint and a low-rank constraint on them, respectively. The sparse tensor is designed to model the sparse noise in the corrupted graphs, and the intrinsic tensor aims to recover the true similarity graphs. We summarize the main contributions of this paper as follows:

1) JPLTD provides a unified framework for IMVC. To alleviate the influence of feature noise in high-dimensional data, JPLTD projects the original data to a lower-dimensional space for compact feature learning, and conducts graph learning based on the reduced dimension space.

2) To reduce the sparse noise interference in complete graph learning, JPLTD introduces the low-rank tensor constraint and decomposes the original tensor into a sparse tensor and an intrinsic tensor, and aims to recover a global low-rank intrinsic tensor with inter-view correlations explored for IMVC. 

3) An iterative optimization algorithm is introduced to solve our model effectively. Extensive experiments on several public datasets verify that JPLTD is superior to some state-of-the-art IMVC methods.

The remainder of this paper is organized as follows. Section \textrm{II} provides notations and preliminaries, including the tensor and graph-based IMVC.  The formulation, optimization, and complexity analysis of JPLTD are presented in Section \textrm{III}. Section \textrm{IV} exhibits the experimental results and discussions. Finally, Section \textrm{V} provides a conclusion for this paper.

\section{Notations and Preliminaries}
\subsection{Notations and tensor preliminaries}
In this paper, tensors, matrices, and vectors are written in calligraphy, bold uppercase, and bold lowercase, respectively. For instance, $\mathcal{Q} \in \mathbb{R}^{n_{1}\times n_{2} \times n_{3}}$, $\textbf{Q}\in\mathbb{R}^{n_{1}\times n_{2}}$, $\textbf{q}\in\mathbb{R}^{n_{1}}$ are a third-order tensor, a matrix, and a vector, respectively. The Frobenius norm, $l_{2,1}$-norm, and nuclear norm of a matrix $\textbf{Q}$ are denoted as $\|\textbf{Q}\|_{F} = \sqrt{\sum_{ij}q_{ij}^{2}}$, $\|\textbf{Q}\|_{2,1}=\sum_{j}\sqrt{\sum_{i}q_{ij}^{2}}$, and $\|\textbf{Q}\|_{*} = \sum_{i}\sigma_{i}(\textbf{Q})$, respectively, where $q_{ij}$ is the $(i,j)$-th element of $\textbf{Q}$ and $\sigma_{i}(\textbf{Q})$ is the $i$-th singular value of $\textbf{Q}$. The  $l_{1}$-norm, Frobenius norm, and infinity norm of a tensor $\mathcal{Q}$ are defined as $\|\mathcal{Q}\|_{1} = \sum_{ijk}|q_{ijk}|$, $\|\mathcal{Q}\|_{F} = \sqrt{\sum_{ijk}q_{ijk}^{2}}$, and $\|\mathcal{Q}\|_{\infty} = \max_{ijk}|q_{ijk}|$, respectively, where $q_{ijk}$ is the $(i,j,k)$-th element of $\mathcal{Q}$. The $i$-th frontal, lateral, and horizantal slice of $\mathcal{Q}$ are denoted as $\mathcal{Q}^{(:,:,i)}$, $\mathcal{Q}^{(:,i,:)}$, and $\mathcal{Q}^{(i,:,:)}$, respectively. Specially, we represent $\mathcal{Q}^{(:,:,i)}$ as $\mathcal{Q}^{(i)}$. $\langle \mathcal{Q}_{1}, \mathcal{Q}_{2}\rangle = \sum_{i=1}^{n_{3}}\langle \mathcal{Q}_{1}^{(i)}, \mathcal{Q}_{2}^{(i)}\rangle$ denotes the inner product of $\mathcal{Q}_{1}$ and $\mathcal{Q}_{2}$. We utilize $\tilde{\mathcal{Q}} = fft(\mathcal{Q},[],3)$ to denote the fast Fourier transformation (FFT) of $\mathcal{Q}$ along the third dimension. The inverse FFT operation is denoted as $\mathcal{Q} = ifft(\tilde{\mathcal{Q}},[],3)$. 
\begin{definition}(\textbf{t-SVD}){\rm \cite{46}}
	For any tensor $\mathcal{Q} \in\mathbb{R}^{n_{1}\times n_{2} \times n_{3}}$, its SVD (Singular Value Decomposition) is defined as: $\mathcal{Q} = \mathcal{U} * \mathcal{S}*\mathcal{V}^{T}$, where $\mathcal{U}\in\mathbb{R}^{n_{1}\times n_{1} \times n_{3}}$ and $\mathcal{V}\in\mathbb{R}^{n_{2}\times n_{2} \times n_{3}}$ are both orthogonal tensors, $\mathcal{S}\in\mathbb{R}^{n_{1}\times n_{2} \times n_{3}}$ is an f-diagonal tensor, whose all frontal slices are diagonal matrices. '*' denotes the t-product.
\end{definition}
\begin{definition}(\textbf{tensor nuclear norm}){\rm \cite{47}}
	The nuclear norm of a tensor $\mathcal{Q}$ is defined as the sum of the singular values of entire frontal slices, i.e., $\|\mathcal{Q}\|_{\circledast} = \sum_{t=1}^{n_{3}}\|\tilde{\textbf{Q}}^{(t)}\|_{*} = \sum_{i=1}^{\min\{n_{1},n_{2}\}}\sum_{t=1}^{n_{3}}\sigma_{i}(\tilde{\textbf{Q}}^{(t)})$. 
\end{definition}

We use $\textbf{X}_{o} = \{\textbf{X}_{o}^{v}\}_{v=1}^{m}$ to denote an incomplete multi-view dataset with $m$ views. $\{\textbf{X}_{o}^{v}\}\in\mathbb{R}^{d_{v}\times n_{v}}$ is the matrix of the corresponding unmissing instances of the $v$-th view, where $d_{v}$ denotes the feature dimension and $n_{v}$ denotes the number of samples in the $v$-th view. For ease of expression, we define $\textbf{X} = \{\textbf{X}^{v}\}_{v=1}^{m}$ as the dataset by completing the missing values of $\textbf{X}_{o}$. $\{\textbf{X}^{v}\}\in\mathbb{R}^{d_{v}\times n}$ corresponds to the matrix of  the $v$-th view, where $n$ represents the number of samples and $n\geq n_{v}$. 

\subsection{Graph-based IMVC preliminaries} 
The graph-based IMVC methods have attracted great attention recently, which attempt to construct the similarity graph to characterize the pairwise similarity among different data points for clustering. Let $\textbf{A}^{v}\in\mathbb{R}^{n_{v}\times n}$ be an index matrix corresponding to the $v$-th view, whose elements are defined as follows:
\begin{eqnarray}
a_{ij}^{v} = 
\left\{
\begin{array}{ll}
1, &\text{if\ the\  $i$-th\ existing\ sample\ in}\ \textbf{X}_{o}^{v}\\ 
 &\text{is the $j$-th sample in}\ \textbf{X}^{v},\\
0, &\text{otherwise}.
\end{array}
\right.
\label{eq1}
\end{eqnarray}
Then, the typical graph-based IMVC framework can be divided into two parts, i.e., graph learning and 
graph fusion\cite{30}:
\begin{equation}\label{eq2}
\begin{aligned}
&\min\limits_{\textbf{G}^{v},\ \textbf{E}^{v}}\|\textbf{E}^{v}\|_{F}^{2} + \lambda \phi(\textbf{G}^{v})\\
&s.t.\ \textbf{E}^{v} = \textbf{X}_{o}^{v} - \textbf{X}_{o}^{v}\textbf{A}^{v}\textbf{G}^{v}\textbf{A}^{v^{T}},
\end{aligned}
\end{equation}
where $\textbf{G}^{v}\in\mathbb{R}^{n\times n}$ denotes the complete graph of the $v$-th view to learn, $\textbf{A}^{v}\textbf{G}^{v}\textbf{A}^{v^{T}}\in\mathbb{R}^{n_{v}\times n_{v}}$ denotes the subgraph corresponding to the $v$-th view, which extracts the subgraph corresponding to existing samples. $\textbf{E}^{v}\in\mathbb{R}^{d_{v}\times n_{v}}$ is the residual matrix, $\phi(\cdot)$ is a norm regularization term about $\textbf{G}^{v}$,  and $\lambda > 0 $ is a balanced parameter.
 It can be easily verified (\ref{eq2}) follows from self-representation
subspace clustering on existing data of $v$-th view. After learning the complete graphs $\{\textbf{G}^{v}\}_{v=1}^{m}$ corresponding to different views, an optimal consensus graph $\textbf{G}\in\mathbb{R}^{n\times n}$ can be obtained by fusing $\{\textbf{G}^{v}\}_{v=1}^{m}$. 
\begin{equation}\label{eq3}
\begin{aligned}
&\min\limits_{\textbf{G}^{v},\ \textbf{E}^{v},\ \textbf{G}}\ \sum_{v=1}^{m}\|\textbf{E}^{v}\|_{F}^{2} + \lambda \Phi(\textbf{G}^{v},\ \textbf{G})\\
&s.t.\ \textbf{E}^{v} = \textbf{X}_{o}^{v} - \textbf{X}_{o}^{v}\textbf{A}^{v}\textbf{G}^{v}\textbf{A}^{v^{T}},
\end{aligned}
\end{equation}
where $\Phi(\cdot)$ is a graph fusion term. The clustering results can be obtained by performing spectral clustering on $\textbf{G}$.

The graph-based approaches have achieved good performance in dealing with MVC and IMVC problems. However, the two-step graph learning methods have some common limitations.  Firstly, many approaches directly use the original high-dimensional data for feature learning, ignoring the noise or outliers in them. Besides, most of them lack the exploration of information among different views. Moreover, due to their lack of consideration of sparse noise in incomplete graphs, their clustering performance is sensitive to the quality of the initial graph.

\section{Methodology}
\begin{figure*}[!t]
	\centering
	\includegraphics[width=18cm,height=5.8cm]{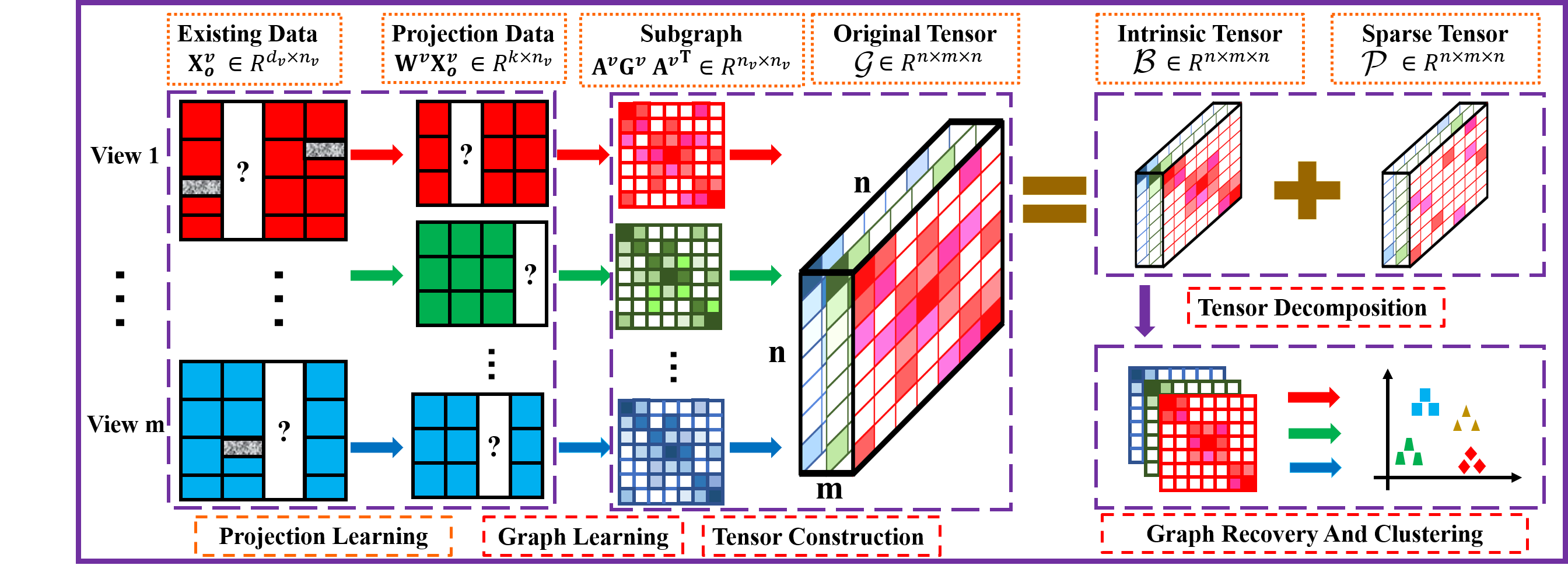}
	\caption{The framework of JPLTD. JPLTD projects the original data into a lower-dimensional space for compact feature learning, and utilizes tensor decomposition to alleviate the influence of the noise introduced by subgraph learning. Finally, JPLTD performs spectral clustering on recovered graphs to obtain the clustering results.}
	\label{fig2}
\end{figure*}

In this section, the formulation and optimization of the proposed Joint Projection learning and Tensor Decomposition (JPLTD) based incomplete multi-view clustering method are introduced in detail. Besides, we analyze the computational complexity of JPLTD. The whole framework of JPLTD is exhibited in Fig.~\ref{fig2}. For conducting compact feature learning, JPLTD introduces projection matrices to project the original data into a lower-dimensional latent space to reduce the influence of original feature redundancy and noise, and conducts graph learning based on this space. To deal with the sparse noise introduced by subgraph learning and tensor completion, JPLTD divides the original tensor into an intrinsic tensor and a sparse noise tensor, which are used to characterize the true data similarities and the noise information, respectively.

\subsection{Model Formulation}
\textit{1) Graph Construction with Projection learning}

As mentioned above, most IMVC methods directly use the original high-dimensional samples and neglect the noise and even outliers interference in them, which is harmful to clustering. Besides, the computation cost of sparse and low-rank representations is very large when the dimension of the features is high\cite{50}. To address these problems, some spatial projection-based approaches are proposed\cite{48,50,51,52,recommand4}. Motivated by them, we introduce projection matrices $\{\textbf{W}^{v}\}_{v=1}^{m}\in\mathbb{R}^{k\times d_{v}}$ to project the original data into a lower-dimensional space for compact feature learning. Simultaneously, we conduct graph learning based on the low-dimension space, which can be formulated as follows:
\begin{equation}\label{eq4}
\begin{aligned}
&\min\limits_{\textbf{G}^{v},\ \textbf{E}^{v},\ \textbf{W}^{v}}\ \sum_{v=1}^{m}\|\textbf{E}^{v}\|_{F}^{2} + \phi(\textbf{G}^{v})\\
s.t.\ \textbf{E}^{v} = &\textbf{W}^{v}\textbf{X}_{o}^{v} - \textbf{W}^{v}\textbf{X}_{o}^{v}\textbf{A}^{v}\textbf{G}^{v}\textbf{A}^{v^{T}},\  \textbf{W}^{v}\textbf{W}^{v^{T}} = \textbf{I}_{k},
\end{aligned}
\end{equation}
where $k$ is the dimension of the obtained latent space, $\{\textbf{E}^{v}\}_{v=1}^{m}\in\mathbb{R}^{k \times n_{v}}$ are residual matrices, $\phi(G^v)$ is a graph regularization term,  and $\textbf{I}_{k}$ is a $k$-order identity matrix. We make the rows of $\{\textbf{W}^{v}\}_{v=1}^{m}$ to be orthogonal, which can keep the solution from being degenerate and improve the optimization efficiency of the model\cite{50}. Besides, since the $l_{2,1}$-norm is more robust to outliers and sample-specific corruptions than the Frobenius norm\cite{48,53}, we introduce the $l_{2,1}$-norm to describe the reconstruction error and change Eq.~(\ref{eq4}) to the following model:
\begin{equation}\label{eq5}
\begin{aligned}
&\min\limits_{\textbf{G}^{v},\ \textbf{E}^{v},\ \textbf{W}^{v}}\ \sum_{v=1}^{m}\|\textbf{E}^{v}\|_{2,1} + \phi(\textbf{G}^{v})\\
s.t.\ \textbf{E}^{v} = &\textbf{W}^{v}\textbf{X}_{o}^{v} - \textbf{W}^{v}\textbf{X}_{o}^{v}\textbf{A}^{v}\textbf{G}^{v}\textbf{A}^{v^{T}},\  \textbf{W}^{v}\textbf{W}^{v^{T}} = \textbf{I}_{k},
\end{aligned}
\end{equation}

\textit{2) Intrinsic Graph Recovery with Tensor Decomposition}

It should be noted that Eq.~(\ref{eq5}) concentrates on the intra-view information and lacks consideration of the inter-view information. Besides, although the learned graphs characterize the sample similarities corresponding to different views, they should have a similar structure, since the instances from all views describe the same object. Therefore, we introduce a tensor nuclear norm  regularization term to explore the correlations across different views. Specifically, we stack the graphs $\{\textbf{G}^{v}\}_{v=1}^{m}$ into a tensor $\mathcal{G}\in\mathbb{R}^{n\times m\times n}$ and use tensor learning to investigate the high-order relationships among all views, and utilize the tenor nuclear norm to characterize the low-rank property of $\mathcal{G}$. By introducing the tensor nuclear norm  regularization term, problem~(\ref{eq5}) is converted into the following form:
\begin{equation}\label{eq6}
\begin{aligned}
&\min\limits_{\textbf{G}^{v},\ \textbf{E}^{v},\ \textbf{W}^{v},\ \mathcal{G}}\ \sum_{v=1}^{m}\|\textbf{E}^{v}\|_{2,1} + \lambda \|\mathcal{G}\|_{\circledast}\\
&s.t.\ \textbf{E}^{v} = \textbf{W}^{v}\textbf{X}_{o}^{v} - \textbf{W}^{v}\textbf{X}_{o}^{v}\textbf{A}^{v}\textbf{G}^{v}\textbf{A}^{v^{T}},\\  &\textbf{W}^{v}\textbf{W}^{v^{T}} = \textbf{I}_{k},\ \mathcal{G} = \Psi(\textbf{G}^{1},\ \textbf{G}^{2},\cdots,\ \textbf{G}^{m}),
\end{aligned}
\end{equation}
where $\Psi(\cdot)$ is an operator to stack the graphs $\{\textbf{G}^{v}\}_{v=1}^{m}$ into a third-order tensor. We can observe that model~(\ref{eq6}) has considered the noise hidden in the original data and the relationship across different views, however, it directly constrains the noisy tensor $\mathcal{G}$. As we illustrated in Fig.~\ref{fig1}, the incomplete graph corresponding to the incomplete data is a subgraph of the complete graph, whose intra-class and inter-class structure have changed. These changes have brought noise information in the obtained graphs which is harmful to clustering. Besides, the missing values in complete graph $\{\textbf{G}^{v}\}_{v=1}^{m}$ of each view  are generally filled by zero, therefore, we regard the harmful information hidden in $\mathcal{G}$ as sparse noise. To alleviate the noise interference, we decompose $\mathcal{G}$ as follows:
\begin{equation}\label{eq7}
\mathcal{G} = \mathcal{B}+\mathcal{P},
\end{equation}  
where $\mathcal{B}\in\mathbb{R}^{n\times m\times n}$ is an intrinsic tensor and $\mathcal{P}\in\mathbb{R}^{n\times m\times n}$ is a sparse noise tensor. The intrinsic tensor $\mathcal{B}$ is designed to depict the true correlations among all instances including missing and existing ones of all views, which satisfies the low-rank constraint. The sparse noise tensor $\mathcal{P}$ is  used to describe the noise information generated from incomplete subgraph learning and missing values imputation. Then, the final object function of JPLTD is written as:
\begin{equation}\label{eq8}
\begin{aligned}
&\min\limits_{\Lambda}\ \sum_{v=1}^{m}\|\textbf{E}^{v}\|_{2,1} + \lambda \|\mathcal{B}\|_{\circledast}+ \theta\|\mathcal{P}\|_{1}\\
s.t.\ &\textbf{E}^{v} = \textbf{W}^{v}\textbf{X}_{o}^{v} - \textbf{W}^{v}\textbf{X}_{o}^{v}\textbf{A}^{v}\textbf{G}^{v}\textbf{A}^{v^{T}},\ \textbf{W}^{v}\textbf{W}^{v^{T}} = \textbf{I}_{k},\\ 
& \mathcal{G} = \mathcal{B}+\mathcal{P},\ \mathcal{G} = \Psi(\textbf{G}^{1},\ \textbf{G}^{2},\cdots,\ \textbf{G}^{m}),
\end{aligned}
\end{equation}
where $\Lambda = \{\textbf{G}^{v},\ \textbf{E}^{v},\ \textbf{W}^{v},\ \mathcal{G},\ \mathcal{B},\ \mathcal{P} \}$ is the variables set to be learned, $\lambda$ and $\theta$ are two balanced parameters. After solving Eq.~(\ref{eq8}), we can obtain the optimal intrinsic tensor $\mathcal{B}_{\dagger}$, and the final similarity matrix $\textbf{H}\in\mathbb{R}^{n\times n}$ can be calculated by:
\begin{equation}\label{eq9}
\textbf{H} = \frac{1}{m}\sum_{v=1}^{m}\frac{|\textbf{B}_{\dagger}^{v}|+ |\textbf{B}_{\dagger}^{v}|^{T}}{2},
\end{equation}
where $\textbf{B}_{\dagger}^{v}$ represents the $v$-th lateral slice of $\mathcal{B}_{\dagger}$ and it also represents the $v$-th graph. We can obtain the final clustering results by performing spectral clustering on \textbf{H}~\cite{48}. 

\subsection{Optimization of JPLTD}
In this subsection, we adopt the alternation direction method of multipliers algorithm (ADMM) to solve the JPLTD model effectively~\cite{54,55,56}. 
To solve (\ref{eq8}), the augmented Lagrangian function is:
 \begin{equation}
\begin{aligned}
L_{\rho} = &\sum_{v=1}^{m}\|\textbf{E}^{v}\|_{2,1} + \lambda \|\mathcal{B}\|_{\circledast}+ \theta\|\mathcal{P}\|_{1}\\
&+\langle\textbf{W}^{v}\textbf{X}_{o}^{v} - \textbf{W}^{v}\textbf{X}_{o}^{v}\textbf{A}^{v}\textbf{G}^{v}\textbf{A}^{v^{T}}-\textbf{E}^{v}, \textbf{J}_{1}^{v}\rangle
+\langle\mathcal{G}-\mathcal{B}-\mathcal{P}, \mathcal{J}_{2}\rangle \\
&+\frac{\rho}{2}\Big(\|\textbf{W}^{v}\textbf{X}_{o}^{v} - \textbf{W}^{v}\textbf{X}_{o}^{v}\textbf{A}^{v}\textbf{G}^{v}\textbf{A}^{v^{T}}-\textbf{E}^{v}\|_{F}^{2}
+\|\mathcal{G}-\mathcal{B}-\mathcal{P}\|_{F}^{2}\Big)\\
s.t.\ &\textbf{W}^{v}\textbf{W}^{v^{T}} = \textbf{I}_{k},\ \mathcal{G} = \Psi(\textbf{G}^{1},\ \textbf{G}^{2},\cdots,\ \textbf{G}^{m}),
\end{aligned}
\label{eq11}
\end{equation}
where $\rho>0$ is a penalty parameter, $\textbf{J}_{1}$ and $\mathcal{J}_{2}$ are Lagrange multipliers. Then, we can solve Eq.~(\ref{eq11}) by optimizing the following subproblems:

\textbf{(1)\ Update $\textbf{W}^{v}$:} Fixing other variables, the minimization problem for updating $\textbf{W}^{v}$ is:
 \begin{equation}
\begin{aligned}
\textbf{W}_{t+1}^{v}=\arg\min\limits_{\textbf{W}^{v}}&\|\textbf{W}^{v}\textbf{X}_{o}^{v} - \textbf{W}^{v}\textbf{X}_{o}^{v}\textbf{A}^{v}\textbf{G}_{t}^{v}\textbf{A}^{v^{T}}-\textbf{E}_{t}^{v}+\frac{\textbf{J}_{1,t}^{v}}{\rho_{t}}\|_{F}^{2}\\
&s.t.\ \textbf{W}^{v}\textbf{W}^{v^{T}} = \textbf{I}_{k},
\end{aligned}
\label{eq12}
\end{equation}
Let $\textbf{F}_{1,t}^{v}$ = $\textbf{X}_{o}^{v} - \textbf{X}_{o}^{v}\textbf{A}^{v}\textbf{G}_{t}^{v}\textbf{A}^{v^{T}}$ and $\textbf{F}_{2,t}^{v}$ = $\textbf{E}_{t}^{v}-\frac{\textbf{J}_{1,t}^{v}}{\rho_{t}}$, then problem~(\ref{eq12}) can be transformed as:
 \begin{equation}
\begin{aligned}
\textbf{W}_{t+1}^{v}&=\arg\min\limits_{\textbf{W}^{v}}\|\textbf{F}_{2,t}^{v}-\textbf{W}^{v}\textbf{F}_{1,t}^{v}\|_{F}^{2}\\
&s.t.\ \textbf{W}^{v}\textbf{W}^{v^{T}} = \textbf{I}_{k},
\end{aligned}
\label{eq13}
\end{equation}
The optimum of $\textbf{W}_{t+1}^{v}$ is provided by $\textbf{W}_{t+1}^{v}$ = $\textbf{V}\textbf{U}^{T}$, where \textbf{U} and \textbf{V} come from the SVD of $\textbf{F}_{1,t}\textbf{F}_{2,t}^{T}$, $i.e.$\ $\textbf{F}_{1,t}\textbf{F}_{2,t}^{T}$ = $\textbf{U}\Sigma\textbf{V}^{T}$.

\textbf{(2)\ Update $\textbf{G}^{v}$:} Fixing other variables, the minimization problem for $\textbf{G}^{v}$ is:
 \begin{equation}
\textbf{G}_{t+1}^{v}=\arg\min\limits_{\textbf{G}^{v}}\|\textbf{F}_{3,t}^{v} - \textbf{W}_{t+1}^{v}\textbf{X}_{o}^{v}\textbf{A}^{v}\textbf{G}^{v}\textbf{A}^{v^{T}}\|_{F}^{2}
+\|\textbf{G}^{v} - \textbf{F}_{4,t}^{v}\|_{F}^{2},
\label{eq14}
\end{equation}
where $\textbf{F}_{3,t}^{v}$ = $\textbf{W}_{t+1}^{v}\textbf{X}_{o}^{v}-\textbf{E}_{t}^{v}+\textbf{J}_{1,t}^{v}/\rho_{t}$ and $\textbf{F}_{4,t}^{v}$ = $\textbf{B}_{t}^{v}+\textbf{P}_{t}^{v}-\textbf{J}_{2,t}^{v}/\rho_{t}$. By making its derivate over $\textbf{G}^{v}$ equal to zero, we have:
 \begin{equation} \Big(\textbf{W}_{t+1}^{v}\textbf{X}_{o}^{v}\textbf{A}^{v}\Big)^{T}\textbf{W}_{t+1}^{v}\textbf{X}_{o}^{v}\textbf{A}^{v}\textbf{G}^{v}+\textbf{G}^{v}\Big(\textbf{A}^{v^{T}}\textbf{A}^{v}\Big)^{-1} = \textbf{F}_{5,t}^{v},
\label{eq15}
\end{equation}
where $\textbf{F}_{5,t}^{v}$ = $\Big[\Big(\textbf{W}_{t+1}^{v}\textbf{X}_{o}^{v}\textbf{A}^{v}\Big)^{T}\textbf{F}_{3,t}^{v}\textbf{A}^{v}+\textbf{F}_{4,t}^{v}\Big]\Big(\textbf{A}^{v^{T}}\textbf{A}^{v}\Big)^{-1}$. Eq.~(\ref{eq15}) is a typical Sylvester equation and we can use the Bartels-Stewart algorithm to solve it~\cite{48}.

\textbf{(3)\ Update $\textbf{E}^{v}$:} With other variables fixed, the optimization problem for updating $\textbf{E}^{v}$ is:
 \begin{equation}
 \textbf{E}_{t+1}^{v} = \arg\min\limits_{\textbf{E}^{v}}\|\textbf{E}^{v}\|_{2,1}+\frac{\rho_{t}}{2}\|\textbf{E}^{v} - \textbf{F}_{6,t}^{v}\|_{F}^{2},
\label{eq16}
\end{equation}
where $\textbf{F}_{6,t}^{v}$ = $\textbf{W}_{t+1}^{v}\textbf{X}_{o}^{v} - \textbf{W}_{t+1}^{v}\textbf{X}_{o}^{v}\textbf{A}^{v}\textbf{G}_{t+1}^{v}\textbf{A}^{v^{T}}+\textbf{J}_{1,t}^{v}/{\rho_{t}}$. To solve problem (\ref{eq16}), we introduce the following theorem\cite{53}:
\begin{theorem}
	Given a matrix \textbf{D}, suppose the optimal solution of
	\begin{equation}
		\min\limits_{\textbf{L}}\eta\|\textbf{L}\|_{2,1}+\frac{1}{2}\|\textbf{L}-\textbf{D}\|_{F}^{2}
		\nonumber 
	\end{equation}
	is $\textbf{L}^{\dagger}$, then the $c$-th column of $\textbf{L}^{\dagger}$ is 
	\begin{eqnarray}
	\textbf{L}_{(:,c)}^{\dagger} = 
	\left\{
	\begin{array}{ll}
	\frac{\|\textbf{D}_{(:,c)}\|_{2}-\eta}{\textbf{D}_{(:,c)}\|_{2}}\textbf{D}_{(:,c)}, &if\ \|\textbf{D}_{(:,c)}\|_{2}>\eta;\\ 
	0, &otherwise.
	\end{array}
	\right.
    \nonumber 
	\end{eqnarray}
	
\end{theorem}
Then, we can easily obtain the solution of $\textbf{E}^{i}$ based on Theorem 1.

\textbf{(4)\ Update $\mathcal{B}$:} With other variables fixed, the optimization problem for $\mathcal{B}$ is:
 \begin{equation}
\mathcal{B}_{t+1} = \arg\min\limits_{\mathcal{B}}\frac{\lambda}{\rho_{t}}\| \mathcal{B}\|_{\circledast}+\frac{1}{2}\|\mathcal{B}-\mathcal{F}_{7,t}\|_{F}^{2},
\label{eq17}
\end{equation}
where $\mathcal{F}_{7,t}$ = $\mathcal{G}_{t+1}-\mathcal{P}_{t}+\frac{\mathcal{J}_{2,t}}{\rho_{t}}$, (\ref{eq17}) is a typical tensor nuclear norm minimization problem, we can solve it by utilizing the following theorem~\cite{57}:
\begin{theorem}
	Given two tensors $\mathcal{L}$, $\mathcal{D}\in\mathbb{R}^{n_{1}\times n_{2}\times n_{3}}$ and a constant $\eta>0$, the  globally optimal solution of
	 \begin{equation}
	 \min\limits_{\mathcal{L}}\|\mathcal{L}\|_{\circledast}+\frac{\eta}{2}\|\mathcal{L}-\mathcal{D}\|_{F}^{2}
	 \nonumber 
	 \end{equation}
	 can be obtained by the tensor tubal-shrinkage operator:
	  \begin{equation}
       \mathcal{L} = \mathcal{K}_{n_{3}\eta}(\mathcal{D}) = \mathcal{U}*\mathcal{K}_{n_{3}\eta}(\mathcal{S})*\mathcal{V}^{T},
	  \nonumber 
	  \end{equation}
	  where $\mathcal{D} = \mathcal{U}*\mathcal{S}*\mathcal{V}^{T}$ and $\mathcal{K}_{n_{3}\eta}(\mathcal{S}) = \mathcal{S}*\mathcal{H}$. $\mathcal{H}$ is an $n_{1}\times n_{2}\times n_{3}$ f-diagonal tensor and its diagonal element
	  in the Fourier domain is $\mathcal{H}_{f}(i,i,j) = \Big(1-\frac{n_{3}\eta}{\mathcal{S}_{f}^{(i)}(i,i)}\Big)_{+}$.
\end{theorem}

\textbf{(5)\ Update $\mathcal{P}$:} We can update $\mathcal{P}$ by working out the following subproblem:
 \begin{equation}
\mathcal{P}_{t+1} = \arg\min\limits_{\mathcal{P}}\frac{\theta}{\rho_{t}}\|\mathcal{P}\|_{1}+\frac{1}{2}\|\mathcal{P}-\mathcal{F}_{8,t}\|_{F}^{2},
\label{eq20}
\end{equation}
where $\mathcal{F}_{8,t}$ = $\mathcal{G}_{t+1}-\mathcal{B}_{t+1}+\mathcal{J}_{2,t}/\rho_{t}$. The optimal solution of problem~(\ref{eq20}) can by obtained by:
  \begin{equation}
 \textbf{P}_{t+1}^{v} = \mathbb{D}_{\frac{\theta}{\rho_{t}}}(\textbf{F}_{8,t}^{v}),
 \label{eq21}
 \end{equation}
where $\mathbb{D}_{\eta}$ is the shrinkage operator defined as $[\mathbb{D}_{\eta}(\textbf{Q})]_{ij}=sign(q_{ij})\cdot\max\{|q_{ij}|-\eta,\  0\}$.

\textbf{(6)\ Update $\rho$, $\textbf{J}_{1}$, $\mathcal{J}_{2}$:} We update the penalty parameter and Lagrange multipliers by:
 \begin{equation}
\begin{aligned}
&\textbf{J}_{1,t+1}^{v} = \textbf{J}_{1,t}^{v} + \rho_{t}\big(\textbf{W}_{t+1}^{v}\textbf{X}_{o}^{v} - \textbf{W}_{t+1}^{v}\textbf{X}_{o}^{v}\textbf{A}^{v}\textbf{G}_{t+1}^{v}\textbf{A}^{v^{T}}-\textbf{E}_{t+1}^{v}\big),\\
&\mathcal{J}_{2,t+1} = \mathcal{J}_{2,t} + \rho_{t}\big(\mathcal{G}_{t+1}-\mathcal{B}_{t+1}-\mathcal{P}_{t+1}\big),\\
&\rho_{t+1} = \min\{\rho_{max},\ \alpha\rho_{t}\},
\end{aligned}
\label{eq22}
\end{equation}
where $\rho_{max}$ is the upper bound of $\rho$ and $\alpha>1$ is a positive scalar. We define the convergence conditions as follows:
\begin{eqnarray}
max
\left\{
\begin{array}{l}
\|\textbf{W}_{t+1}^{v}\textbf{X}_{o}^{v} - \textbf{W}_{t+1}^{v}\textbf{X}_{o}^{v}\textbf{A}^{v}\textbf{G}_{t+1}^{v}\textbf{A}^{v^{T}}-\textbf{E}_{t+1}^{v}\|_{\infty} \\ 
\|\mathcal{G}_{t+1}-\mathcal{B}_{t+1}-\mathcal{P}_{t+1}\|_{\infty}
\end{array}
\right\}<\epsilon,
\label{eq23}
\end{eqnarray}
where $\epsilon$ is a small tolerance. The overall optimzation procedure of JPLTD is summarized in Algorithm \ref{alg1}.
\begin{algorithm}[t]
	\caption{\textbf{JPLTD Algorithm}}
	\label{alg1}
	\textbf{Input:} Incomplete multi-view data $\{\textbf{X}_{o}^{v}\}_{v=1}^{m}$, model parameters $\lambda$, $\theta$, $k$;\\
	\textbf{Output:} The Final Clustering Results.
	\begin{algorithmic}[1]
		\State Initialize $\textbf{G}^{v}$=0, $\textbf{E}^{v}$=0, $\textbf{J}_{1}^{v}$=0, $\mathcal{B}$=$\mathcal{P}$=0, $\mathcal{J}_{2}$=0,  $\alpha$=1.3, $\rho$ = 1e-3, $\rho_{max}$ = 1e6, $\epsilon$ = 1e-5;
		\While{not converged}
		\State Update $\textbf{W}^{v}$ by solving Eq.~(\ref{eq13});
		\State Update $\textbf{G}^{v}$ by solving Eq.~(\ref{eq15});
		\State Update $\textbf{E}^{v}$ by solving Eq.~(\ref{eq16});
		\State Construct tensor $\mathcal{G}$ = $\Psi\{\textbf{G}^{1},\ \textbf{G}^{2},\cdots,\ \textbf{G}^{m}\}$;
		\State Update $\mathcal{B}$ by Eq.~(\ref{eq17});
		\State Update $\mathcal{P}$ by Eq.~(\ref{eq21});
		\State Update the Lagrange multipliers and penalty factors by Eq.~(\ref{eq22});
		\EndWhile
		\State Compute the final similarity graph $\textbf{H}$ by Eq.~(\ref{eq9}).
		\State Perform spectral clustering on $\textbf{H}$.
		\State Return clustering results.
	\end{algorithmic}
\end{algorithm}
\subsection{Computational Complexity Analysis}
Herein, we discuss the computational time consumption of Algorithm \ref{alg1}. In each iteration, it takes $O(k^{2}d_{v})$ with $k<d_{v}$ to conduct the SVD operation for updating $\textbf{W}^{v}$. The main time costs for updating $\textbf{G}^{v}$ are the Sylvester equation and matrix inverse operation. It should be noted that the index matrix $\textbf{A}^{v}$ is fixed and we can compute $\textbf{A}^{v^{T}}\textbf{A}^{v}$ outside the loop. The time consumption for solving the Sylvester equation is $O(n^{3})$. As for updating $\mathcal{B}$, we mainly consider the computational complexity of FFT, inverse FFT, and SVD operations. For an $n\times m\times n$ tensor, it takes $O(mn^{2}log(n))$
to conduct FFT and inverse FFT operations,  and needs $O(m^{2}n^{2})$ for SVD operation~\cite{30}. Therefore, it needs $O(mn^{2}log(n)+m^{2}n^{2})$ to update $\mathcal{B}$. The updates of the rest variables only have some basic matrix operations, whose computational costs are lower than the above variables. Therefore, if there are $\beta$ iterations, the overall computational complexity  of the update process is about $O(\beta(n^{3}+mn^{2}log(n)+m^{2}n^{2}+k^{2}d_{v}))$. After obtaining \textbf{H}, we perform spectral clustering on \textbf{H} and get clustering results, this process needs O($n^{3}$).

\section{Experiments}
In this section, we utilize some benchmark datasets to conduct experiments and several state-of-the-art IMVC methods for comparison to prove the effectiveness and robustness of JPLTD. Subsequently, the model parameters analysis and the convergence analysis of JPLTD  will be given. 

\subsection{Experimental Settings}
\textit{1) Datasets:} Seven public datasets exhibited in Table \ref{tab1} are utilized in our experiments to evaluate the clustering performance of JPLTD.  
\begin{table}[!t]
	\footnotesize
	\centering
	\caption{datasets in experiments.}
	\label{tab1}
	\renewcommand{\arraystretch}{1.0}
	\tabcolsep 6pt
	\begin{tabular}{ccccc}
		\toprule
		Datasets & size & classes & views & Dimensionality\\
		\midrule
		ORL  & 400 & 40 & 3 & 4096, 3304, 6750\\
		NGs & 500 & 5 & 3 & 2000, 2000, 2000\\
		MSRC-v1 &	210	& 7	& 4	& 24, 512, 256, 254\\			
		BBCSport & 544 & 5 & 2 & 3183, 3203\\
		RGB-D   & 1449 & 11 & 2 & 2048, 300\\
		UCI     & 2000 & 10 & 2 & 240, 76\\
		Scene     & 4485 & 15 & 3 &1800, 1180, 1240\\
		\bottomrule
	\end{tabular}
\end{table}

\begin{itemize}
\item $\textbf{ORL}\footnote{\url{https://cam-orl.co.uk/facedatabase.html}}$ consists of 400 face images from 40 different individuals, each people has 10 samples.  Three types of features (i.e., intensity, LBP, and Gabor) are extracted as three views, whose dimensions are 4096, 3304, and 6750, respectively.
\item $\textbf{NGs}\footnote{\url{http://qwone.com/~jason/20Newsgroups/}}$ is a subset of the 20NG dataset, which is comprised of 500 files with 5 tags. The dataset contains three views with 2000, 2000, and 2000  characteristic dimensions, respectively.
\item $\textbf{MSRC-v1}$~\cite{BBCSport} is a widely used scene dataset, which contains 210 images of 7 natural scene categories, including building, cow, airplane, tree, bicycle, face, and car. Four types of features are extracted as four views, whose dimensions are 24, 512, 256, and 254, respectively. 
\item $\textbf{BBCSport}$~\cite{BBCSport} is comprised of 544 articles on 5 sports topics from the BBC news website. It has two views with 3183 and 3203 dimensions, respectively.
\item $\textbf{RGB-D}$~\cite{RGB-D} is an indoor scene image-text dataset, which consists of 1449 indoor scene images and text descriptions of the images. The characteristic dimensions of the visual and textual views are 2048 and 300, respectively.
\item $\textbf{UCI}\footnote{\url{http://ss.sysu.edu.cn/~py/}}$ has 2000 handwritten digit
images. Two types of features are extracted as two views whose dimensions are 240 and 76, respectively.
\item $\textbf{Scene}$~\cite{57} is comprised of 4485 images of 15 natural scene categories. Three views are formed by extracting three types of features (i.e., GIST, PHOG, and LBP), whose dimensions are 1800, 1180, and 1240, respectively.
\end{itemize}
\begin{table*}[t]
	\scriptsize
	\centering
	\caption{ Mean ACC (\%), NMI (\%), ARI (\%) of different approaches on seven datasets. The best results are highlighted with bold.}
	\label{tab2} 
	\renewcommand{\arraystretch}{0.92}
	\scalebox{0.97}{ 
	\begin{tabular}{|p{0.1cm}|p{1.5cm}|p{0.87cm} p{0.87cm} p{1.0cm} | p{0.87cm} p{0.87cm} p{1.0cm} |p{0.87cm}p{0.87cm} p{1.0cm}|p{0.87cm}p{0.87cm} p{1.0cm}|}
		\hline
		& Missing rates&\multicolumn{3}{c}{0.1}\vline & \multicolumn{3}{c}{0.3}\vline & \multicolumn{3}{c}{0.5}\vline& \multicolumn{3}{c}{0.7}\vline \\
		\hline
		& Metrics  & ACC~(\%)   & NMI~(\%)   & ARI~(\%)   & ACC~(\%)   & NMI~(\%)   & ARI~(\%) &    ACC~(\%)   & NMI~(\%)   & ARI~(\%)   &  ACC~(\%)   & NMI~(\%)   & ARI~(\%) \\
		\hline
		\multirow{10}{*}{\rotatebox{90}{\textbf{ORL}}}
		&UEAF	&63.52±2.84 	&78.45±1.62 	&47.03±3.62 	&50.32±2.64 	&70.34±1.30 	&31.91±2.70 	&45.85±1.60 &	62.65±1.00 	&12.09±1.76 	&42.50±2.33 	&58.16±1.49	 &8.23±2.59\\
		&CLIMC-SC	&80.85±1.58	&90.15±1.06	&73.34±3.14	&80.40±2.19	&89.51±1.09	&72.19±2.97	&75.90±1.51	&85.55±0.71	&64.19±2.01	&71.30±4.10	&82.86±1.93	&58.76±4.63\\
		&IMVC-CBG	&74.02±2.88	&88.42±1.57	&64.43±4.71	&73.00±2.13	&88.50±1.55	&63.88±4.88	&72.68±2.80	&88.42±0.94	&64.01±2.73	&71.09±4.23	&87.85±1.73	&62.98±4.51\\
		&IMVTSC-MVI &84.13±2.50 &94.40±0.45 &80.26±2.31 &81.67±1.99 &90.86±0.67 &74.26±2.25 &80.83±1.76 &90.95±0.70 &73.64±1.88 &78.67±2.48 &89.44±0.86 	&70.54±2.68\\
		&TMBSD	&96.00±1.38 &98.02±0.36 &95.07±1.45 &94.60±1.80 &97.97±0.56 &92.97±2.03 	&92.54±1.71 &96.83±0.88 &89.98±2.00 &91.30±1.63 &96.06±0.67	&87.93±1.81\\
		&TC-IMVC &78.70±1.86 &89.20±0.56 &71.09±1.09	&77.05±2.54	&87.43±0.99 	&65.99±2.04 &75.05±0.54	&85.12±0.51	&63.00±0.95	&73.80±2.24	&84.85±0.84 	&62.28±1.44\\
		&VCL-IMVC	&72.30±1.34	&86.23±0.84	&60.71±1.39	&69.75±1.70	&85.63±0.75	&57.53±1.31	&66.50±2.54	&82.04±0.73	&54.10±2.00	&66.00±1.12	&83.49±0.48	&53.99±1.18\\
		&HCP-IMSC	&82.08±2.02	&91.29±0.67	&75.65±2.14	&81.63±1.58	&91.52±0.57	&75.51±1.36	&80.80±1.40	&90.21±0.67	&72.81±2.00	&81.51±0.84	&88.87±0.68	&71.97±0.92\\
		&\textbf{JPLTD}	&\textbf{97.65±1.72}&\textbf{99.47±0.38}&\textbf{97.50±1.73}&\textbf{97.51±1.50} &\textbf{99.43±0.32}&\textbf{97.46±1.51}&\textbf{96.36±1.33}&\textbf{98.68±0.32}&\textbf{95.29±1.36}&\textbf{95.66±1.72}&\textbf{98.62±0.45}&\textbf{95.03±1.76}\\
		\hline
		\multirow{10}{*}{\rotatebox{90}{\textbf{NGs}}}
		&UEAF	&94.20±0.00	&83.61±0.00	&86.02±0.00	&92.80±0.00	&80.95±0.00	&82.78±0.00	&87.00±0.00	&69.93±0.00	&70.43±0.00	&76.10±0.32	&56.58±0.44	&51.54±0.47\\
		&CLIMC-SC	&97.00±0.00	&90.98±0.00	&92.66±0.00	&93.20±0.00	&81.72±0.00	&83.71±0.00	&90.20±0.00	&74.16±0.00	&77.06±0.00	&87.20±0.00	&66.68±0.00	&70.42±0.00\\
		&IMVC-CBG	&90.60±0.00	&78.35±0.00	&78.11±0.00	&89.26±0.09	&74.37±0.15	&74.88±0.21	&84.42±0.11	&66.36±0.14	&63.80±0.23	&80.44±0.08	&59.99±0.24	&55.72±0.15\\
		&IMVTSC-MVI &99.40±0.00 &98.08±0.00	&98.50±0.00	&99.20±0.00	&97.39±0.00	&98.00±0.00	&99.20±0.00	&97.22±0.00	&98.00±0.00	&99.00±0.00	&96.52±0.00	&97.50±0.00\\
		&TMBSD	&99.20±0.00	&97.21±0.00	&98.00±0.00	&99.00±0.00	&96.86±0.00	&97.51±0.00	&99.00±0.00	&96.69±0.00	&97.50±0.00	&98.60±0.00	&95.47±0.00	&96.50±0.00\\
		&TC-IMVC	&96.20±0.00	&88.49±0.00	&90.72±0.00	&92.20±0.00	&79.36±0.00	&81.54±0.00
		&89.60±0.00	&73.51±0.00	&75.83±0.00 &83.20±0.00	&60.64±0.00	&62.95±0.00\\
		&VCL-IMVC	&92.80±0.00	&79.96±0.00	&82.89±0.00	&91.00±0.00	&75.89±0.00	&78.79±0.00	&86.4±0.00	&64.67±0.00	&68.88±0.00	&84.40±0.00	&60.64±0.00	&64.68±0.00\\
		&HCP-IMSC	&97.60±0.00	&92.92±0.00	&94.08±0.00	&94.20±0.00	&84.46±0.00	&86.06±0.00	&89.60±0.00	&73.42±0.00	&75.60±0.00	&86.20±0.00	&65.77±0.00	&68.46±0.00\\
		&\textbf{JPLTD}	&\textbf{100.0±0.00}	&\textbf{100.0±0.00}	&\textbf{100.0±0.00}	&\textbf{99.80±0.00}	&\textbf{99.30±0.00}	&\textbf{99.50±0.00}	&\textbf{99.60±0.00}	&\textbf{98.61±0.00}	&\textbf{98.99±0.00}	&\textbf{99.60±0.00}	&\textbf{98.61±0.00}	&\textbf{98.99±0.00}\\
		\hline
		\multirow{10}{*}{\rotatebox{90}{\textbf{MSRC-v1}}}
		&UEAF	&61.92±0.33	&70.71±0.25	&60.36±0.31	&57.21±0.85	&66.29±0.38	&46.07±0.01	&38.20±0.67	&51.62±0.72	&37.92±0.98	&27.05±0.55	&41.43±0.84	&28.11±0.47
		\\
		&CLIMC-SC	&77.71±1.90	&68.64±1.13	&61.14±2.00	&75.71±2.55	&68.51±1.84	&60.36±2.96	&74.39±1.86	&66.21±2.30	&57.44±2.82	&74.10±1.66	&62.21±2.11	&53.13±1.80
		\\
		&IMVC-CBG	&64.45±0.02	&55.15±0.02	&42.13±0.03	&63.81±0.03	&53.77±0.03	&41.48±0.04	&62.67±0.04	&54.17±0.02	&40.88±0.03	&62.57±0.03	&53.89±0.03	&40.70±0.03
		\\
		&IMVTSC-MVI &95.24±0.00	&90.77±0.00	&89.26±0.00	&93.90±0.00	&88.23±0.01	&86.80±0.02	&93.81±0.00	&87.94±0.16	&86.20±0.12	&81.14±0.00	&73.69±0.01	&66.41±0.00
		\\
		&TMBSD	&99.52±0.00	&98.92±0.00	&98.88±0.00	&99.05±0.00	&97.85±0.00	&97.77±0.00	&98.10±0.00	&96.04±0.00	&95.60±0.00	&98.10±0.00	&96.03±0.00	&95.54±0.00
		\\
		&TC-IMVC	&76.71±0.42	&69.67±0.47	&60.89±0.46	&76.67±0.55	&68.27±0.75	&58.55±1.12	&72.81±0.35	&64.42±0.62	&53.42±0.72	&72.05±0.39	&61.76±0.59	&53.11±0.75
		\\
		&VCL-IMVC	&73.81±0.00	&66.95±0.09	&58.35±0.10	&72.95±0.52	&62.54±1.87	&52.61±1.74	&72.28±0.62	&59.53±0.47	&51.20±0.51	&69.14±0.21	&59.09±0.48	&50.36±0.44
		\\		
		&HCP-IMSC	&84.07±0.29	&72.87±0.32	&67.15±0.45	&80.48±0.00	&66.21±0.00	&60.04±0.00	&77.67±0.41	&62.90±0.61	&57.24±0.71	&72.64±0.24	&58.40±0.21	&50.10±0.20
		\\	
		&\textbf{JPLTD}	&\textbf{100.0±0.00
		} &\textbf{100.0±0.00}	&\textbf{100.0±0.00}	&\textbf{99.52±0.00}	&\textbf{98.92±0.00}	&\textbf{98.88±0.00}	&\textbf{99.28±0.24}	&\textbf{98.38±0.54}	&\textbf{98.31±0.57}	&\textbf{99.27±0.24}	&\textbf{98.36±0.54}	&\textbf{98.30±0.56}\\
		\hline
		\multirow{10}{*}{\rotatebox{90}{\textbf{BBCSport}}}
		&UEAF	&84.38±2.01 &68.67±0.72	&66.15±1.25	&84.45±0.37 &65.03±1.19 &65.92±1.16	&81.49±0.93   &63.43±1.82	&64.69±0.68	&77.76±1.03  &58.06±1.32 &54.17±1.07\\
		&CLIMC-SC	&87.79±0.51	&79.24±1.40	&81.38±1.21	&86.62±0.33	&76.10±0.93	&78.88±1.10	&86.40±0.49	&76.22±1.60	&78.44±1.44	&85.18±0.79	&72.44±1.69	&74.68±1.99\\
		&IMVC-CBG	&88.60±1.76 &73.14±1.35 &74.57±2.69 &88.10±0.91 &72.35±2.24 	&74.17±2.16 &88.11±2.11 &72.23±1.61 &72.21±3.24 &86.03±0.99 &69.18±0.55 	&68.18±0.89\\
		&IMVTSC-MVI &96.27±1.88  &94.46±0.58 &94.45±1.96 &96.08±1.91 &94.40±1.38 	&94.23±2.81	&94.92±0.97 &92.37±1.84 &93.32±0.56 &91.17±1.87	&87.29±1.41 	&89.54±2.82\\
		&TMBSD	&94.04±0.97	&85.77±2.22	&87.23±1.57	&92.66±2.06 &84.55±2.02 &82.52±1.54 	&91.54±2.13 &84.00±1.04 &80.12±1.46	 &88.05±0.55 &83.64±1.01 &73.98±1.27\\
		&TC-IMVC	&85.08±1.12 &75.63±2.22 &72.03±1.10	 &82.50±1.74 &71.00±0.51 	&69.55±1.68 &80.15±1.63 &70.18±1.23 &64.00±2.42 &78.13±0.87 &66.22±1.29 	&64.29±1.83\\
		&VCL-IMVC	&93.38±0.20	&81.53±0.13	&84.53±0.12	&90.26±0.10	&73.72±0.12	&76.94±0.23	&86.40±0.39	&72.55±0.72	&69.13±0.24	&79.41±0.10	&66.75±0.07	&68.17±0.17\\		
		&HCP-IMSC	&94.30±0.00	&83.52±0.38 &85.72±0.00 &93.57±0.00 &81.51±0.00	&83.75±0.00	&94.67±0.00	&84.13±0.00	&86.48±0.00	&90.46±0.05	&74.25±0.12	&75.64±0.15\\	
		&\textbf{JPLTD}	&\textbf{99.91±0.09} &\textbf{99.67±0.34}	&\textbf{99.83±0.17}	&\textbf{99.63±0.19}	&\textbf{98.69±0.70}	&\textbf{99.32±0.45}	&\textbf{99.45±0.19}	&\textbf{98.13±0.57}	&\textbf{98.76±0.55}	&\textbf{98.90±0.37}	&\textbf{96.50±0.92}	&\textbf{97.30±0.96}\\
		\hline
		\multirow{10}{*}{\rotatebox{90}{\textbf{RGB-D}}}
		&UEAF	&44.29±0.93 &37.47±1.31 &25.44±2.38 &38.16±1.87	&32.21±1.08 &18.23±1.92 	&35.97±1.81 &29.86±1.37 &13.38±0.83 &33.18±1.99 &26.80±2.67 &9.09±1.20\\
		&CLIMC-SC	&48.20±0.64	&38.83±1.07	&27.58±0.91	&43.41±0.94	&33.39±0.93	&22.36±0.68	&38.54±0.94	&28.04±1.02	&18.21±0.51	&32.93±1.95	&24.88±0.81	&13.78±1.22\\
		&IMVC-CBG	 &47.73±3.65 &34.84±2.63 &26.38±2.82 &46.72±2.66 &34.15±2.23 	&25.66±2.46 &45.69±3.69 &33.96±0.77 &24.56±3.52 &44.81±2.27 &34.05±0.73 	&24.33±1.70\\
		&IMVTSC-MVI &44.28±1.94 &51.06±0.91 &28.12±2.28 &40.60±1.07 &32.16±0.60 &19.37±0.78 &34.74±0.34 &27.43±0.25 &13.58±0.33 &32.62±0.26 &23.14±0.16 &12.67±0.31\\
		&TMBSD	&42.84±0.32 &34.61±0.30 &21.81±1.23 &37.68±0.86 &27.57±0.59 &17.56±0.64 	&32.65±0.21 &23.76±0.15 &14.67±0.22 &32.74±1.23 &21.43±1.09 &13.12±0.96\\
		&TC-IMVC	&49.36±0.92 &42.30±0.96 &32.68±0.91 &41.66±0.98 &35.84±1.34 	&20.64±1.01 &38.62±0.84 &33.20±1.23	 &16.07±1.48 &34.64±1.92 &30.55±1.44 	&10.57±1.25\\
		&VCL-IMVC	&42.97±0.94	&30.35±0.40	&22.17±0.42	&38.76±0.68	&27.42±0.38	&17.90±0.38	&36.41±0.84	&22.16±0.31	&14.64±0.40	&32.04±0.98	&20.05±0.41	&12.52±0.40\\
		&HCP-IMSC	&47.40±0.46	&38.62±0.25	&27.21±0.30	&45.48±0.50	&34.54±0.56	&25.37±0.76	&41.78±0.38	&31.56±0.16	&21.22±0.63	&37.37±0.53	&28.60±0.37	&18.14±0.35\\
		&\textbf{JPLTD}	&\textbf{64.73±0.04}	&\textbf{62.26±0.10}	&\textbf{50.10±0.10}	&\textbf{62.52±0.08}	&\textbf{59.65±0.13}	&\textbf{46.16±0.12}	&\textbf{62.29±0.06}	&\textbf{59.25±0.08}	&\textbf{45.45±0.08}	&\textbf{61.70±0.33}	&\textbf{55.23±0.18}	&\textbf{43.72±0.19}\\
		\hline
		\multirow{10}{*}{\rotatebox{90}{\textbf{UCI}}}
		&UEAF	&87.40±1.66	&78.20±0.68	&74.49±0.73	&68.62±0.81	&62.35±1.68	&50.55±1.76	&56.85±0.63	&51.17±1.23	&32.18±0.73	&43.92±1.35	&41.98±1.26	&17.19±1.92\\
		&CLIMC-SC	&96.57±0.21	&92.71±0.44	&92.50±0.46	&90.69±4.66	&86.82±1.20	&83.76±3.26	&88.71±1.39	&83.97±0.41	&79.31±1.24	&88.04±5.95	&87.93±1.04	&83.27±4.73\\
		&IMVC-CBG	&79.54±1.23	&72.62±0.74	&65.57±0.66	&79.19±1.05	&72.38±1.32	&65.43±1.43	&76.20±0.77	&71.29±1.16	&63.21±0.91	&71.49±1.84	&66.68±0.79	&56.90±1.27\\
		&IMVTSC-MVI&99.60±0.24	&99.01±0.37	&99.11±0.35	&99.35±0.27	&98.32±0.24	&98.56±0.16	&98.90±0.37	&97.05±0.68	&97.56±0.42	&97.80±0.94	&94.63±1.27	&95.18±0.98\\
		&TMBSD	&99.15±0.28	&98.00±0.35	&98.13±0.38	&98.98±0.29	&97.60±0.28	&97.77±0.61	&98.05±0.40	&95.85±1.44	&95.72±0.91	&96.11±1.17	&94.41±0.98	&92.24±1.33\\
		&TC-IMVC	&81.25±1.24	&82.62±1.59	&75.30±1.87	&73.37±1.80	&78.42±1.52	&69.06±2.12	&68.90±1.44	&74.99±1.11	&60.44±1.51	&48.88±1.34	&45.85±1.60	&22.65±1.16\\
		&VCL-IMVC	&91.59±0.06	&87.33±0.08	&83.73±0.11	&89.67±0.77	&84.60±0.26	&80.54±1.07	&89.10±0.84	&83.64±0.43	&79.70±1.39	&88.73±1.49	&82.40±2.19	&78.78±2.83\\
		&HCP-IMSC	&88.75±0.05	&80.72±0.05	&77.74±0.08	&80.20±0.00	&77.00±0.00	&70.50±0.00	&79.50±0.06	&76.30±0.00	&69.53±0.00	&78.10±0.00	&75.25±0.02	&68.48±0.02\\		
		&\textbf{JPLTD}	&\textbf{99.65±0.05}	&\textbf{99.13±0.04}	&\textbf{99.27±0.04}	&\textbf{99.45±0.16}	&\textbf{98.64±0.04}	&\textbf{98.84±0.28}	&\textbf{99.42±0.03}	&\textbf{98.53±0.08}	&\textbf{98.72±0.03}	&\textbf{99.35±0.06}	&\textbf{98.38±0.39}	&\textbf{98.59±0.18}
		\\
		\hline
		\multirow{10}{*}{\rotatebox{90}{\textbf{Scene}}}
		&UEAF	&40.15±1.85	&45.07±0.57	&26.06±1.94	&36.07±1.89	&37.49±1.34	&19.98±2.82	&29.96±0.97	&31.21±0.83	&15.12±1.34	&22.99±0.94	&23.89±0.68	&7.97±1.30\\
		&CLIMC-SC	&57.37±0.32	&61.14±0.23	&43.90±0.32	&55.79±0.23	&58.57±0.29	&41.63±0.22	&54.43±0.80	&56.05±0.69	&39.65±0.94	&53.63±0.76	&51.93±0.45	&36.32±0.78\\
		&IMVC-CBG	&51.22±1.59	&48.85±1.33	&33.26±1.09	&51.39±2.03	&48.37±1.13	&32.94±1.30	&50.40±1.96	&48.72±0.88	&32.81±1.03	&50.38±2.12	&49.47±0.67	&33.54±0.84\\	
		&IMVTSC-MVI	&76.43±0.66	&77.09±0.71	&67.87±1.79	&66.64±2.53	&64.43±1.76	&52.55±1.72	&59.67±1.32	&56.38±2.64	&41.80±0.81	&52.71±0.95	&49.47±1.60	&31.96±0.94\\
		&TMBSD	&59.90±0.87	&56.93±0.64	&59.59±0.63	&52.33±1.25	&50.79±0.55	&51.17±1.62	&43.06±1.58	&40.88±1.17	&42.59±2.34	&31.57±0.85	&29.61±1.65	&30.82±0.81\\
		&TC-IMVC	&54.83±1.25	&58.88±0.74	&40.63±0.84	&52.12±1.43	&52.41±0.82	&36.85±0.76	&49.53±1.49	&52.69±1.39	&34.06±1.02	&46.37±1.29	&48.14±0.54	&33.25±1.16\\
		&VCL-IMVC	&52.08±0.87	&53.77±0.15	&36.71±0.27	&51.38±0.96	&51.28±0.35	&34.67±0.54	&46.95±0.33	&46.68±0.47	&29.05±0.44	&45.35±0.67	&43.59±0.29	&28.28±0.27\\
		&HCP-IMSC	&48.94±0.07	&50.17±0.35	&33.83±0.39	&47.62±1.35	&49.50±0.18	&32.30±0.85	&46.61±0.52	&47.59±0.23	&30.57±0.26	&40.46±0.13	&41.53±0.06	&24.87±0.09\\
		&\textbf{JPLTD}	&\textbf{87.69±0.04}	&\textbf{90.50±0.07}	&\textbf{84.48±0.09}	&\textbf{87.58±0.01}	&\textbf{90.04±0.03}	&\textbf{84.23±0.03}	&\textbf{86.96±0.02}	&\textbf{89.15±0.04}	&\textbf{83.23±0.04}	&\textbf{86.73±0.02}	&\textbf{88.43±0.03}	&\textbf{82.58±0.02}
		\\
		\hline
	\end{tabular}%
}
\end{table*}%
\begin{figure*}[t]
	\centering
	\begin{minipage}[c]{0.495\textwidth}
		\centering
		\includegraphics[width=9cm,height=3cm]{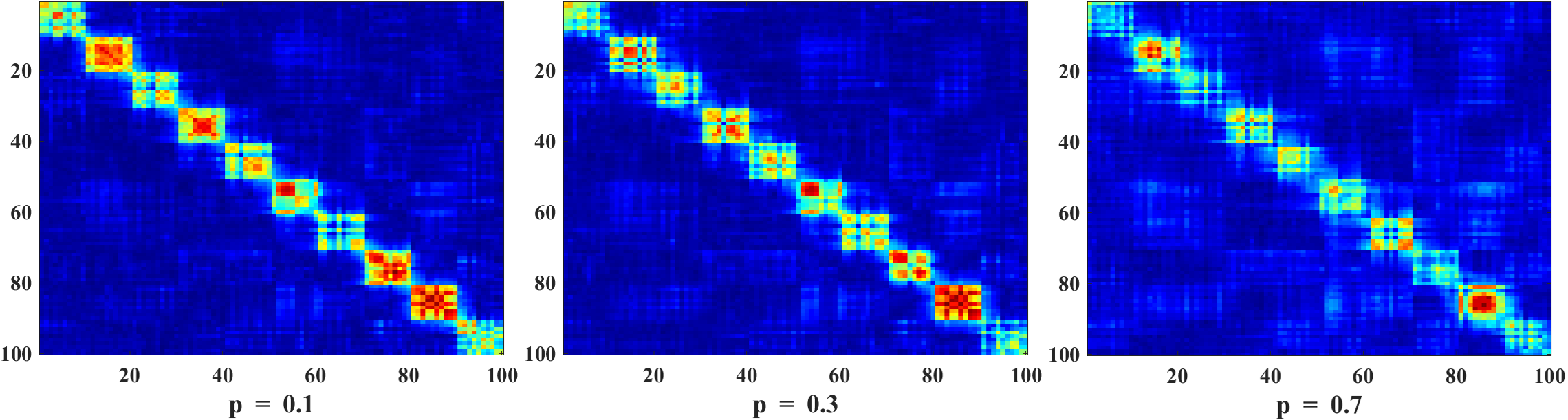}
		\centerline{(a)\ JPLTD on ORL}
	\end{minipage}
	\begin{minipage}[c]{0.495\textwidth}
		\centering
		\includegraphics[width=9cm,height=3cm]{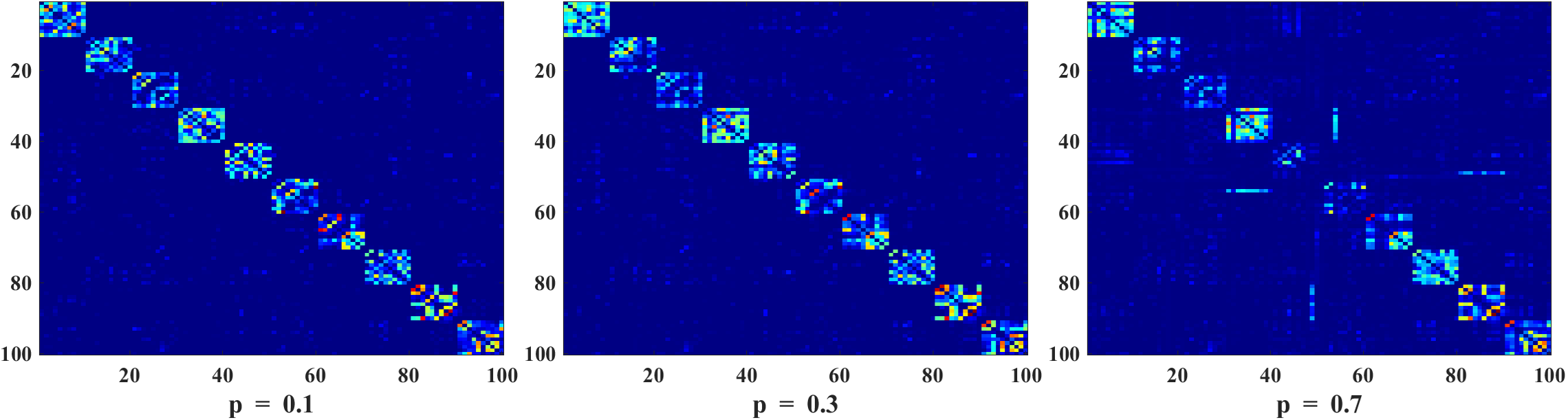}
		\centerline{(b)\ IMVTSC-MVI on ORL}
	\end{minipage}\\[3mm]
	\begin{minipage}[c]{0.495\textwidth}
		\centering
		\includegraphics[width=9cm,height=3cm]{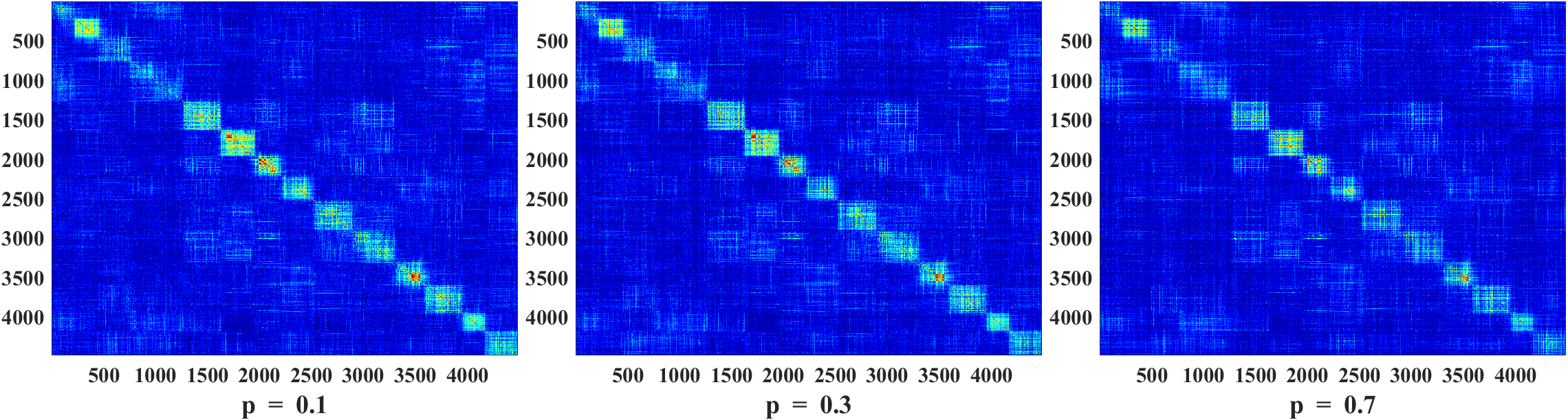}
		\centerline{(c)\ JPLTD on Scene}
	\end{minipage}
	\begin{minipage}[c]{0.495\textwidth}
		\centering
		\includegraphics[width=9cm,height=3cm]{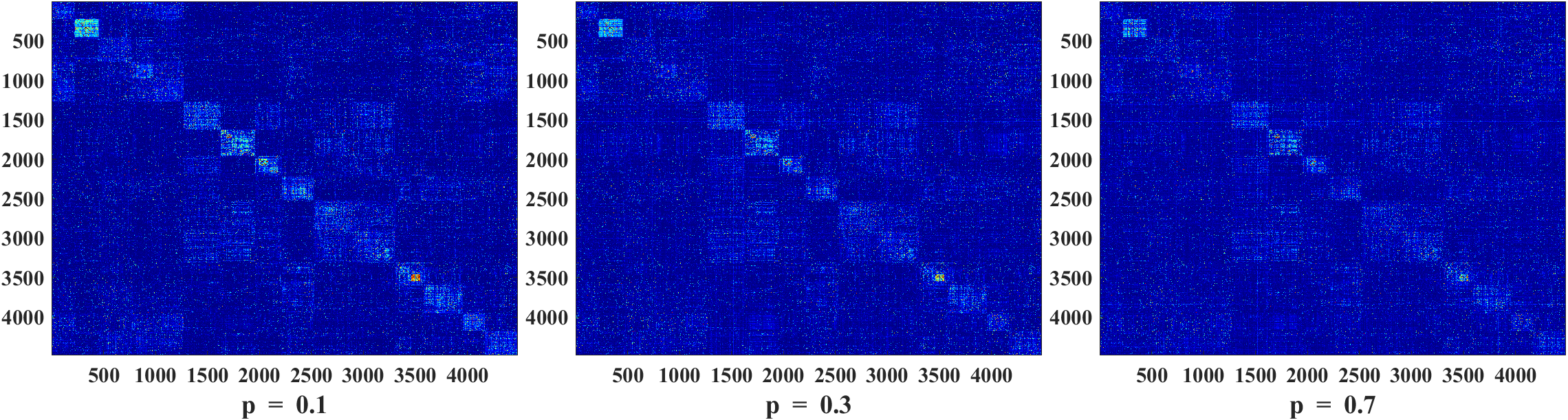}
		\centerline{(d)\ IMVTSC-MVI on Scene}
	\end{minipage}\\[3mm]
	\caption{Visualization of recovered graphs of JPLTD and IMVTSC-MVI on
		ORL and Scene datasets with varying missing rate \textit{p}. }
	\label{fig3}
\end{figure*}

\textit{2) Baselines:} We adopt eight state-of-the-art IMVC approaches to compare the clustering performance of JPLTD, including UEAF~\cite{36}, CLIMC-SC~\cite{38},  IMVC-CBG~\cite{30}, IMVTSC-MVI~\cite{33}, TMBSD~\cite{34}, TC-IMVC~\cite{31}, VCL-IMVC~\cite{45}, and HCP-IMSC~\cite{40}, among which the last five are tensor-based IMVC methods. 
The consensus learning-based incomplete multi-view clustering (CLIMC)~\cite{38} method makes the similarity graph be correlated with the consensus representation by a graph Laplacian regularization. CLIMC-SC performs spectral clustering on the consensus similarity graph of CLIMC. The other approaches have been introduced in Section I.

\textit{3) Evaluation Metrics:} Three popular metrics  i.e., Accuracy (ACC), Normalized Mutual Information (NMI), and Adjusted Rand Index (ARI) are utilized to evaluate the clustering performance. For all criteria, the larger values indicate better performance. We run each experiment 20 times to report the average values and standard deviation.

\textit{4) Incomplete Data Construction:} Similar to \cite{31}, we construct incomplete datasets with four missing rates (0.1, 0.3, 0.5, 0.7). For instance, if the missing rate is 0.3, we randomly remove partial views of the 30\% samples and preserve the other 70\% samples as complete data.

\textit{5) Parameter Settings:} The parameter settings of baselines follow the suggested range in the corresponding literature to report the best results. As for JPLTD, the dimension of projection space $k$,  model parameters $\lambda$ and $\theta$ are selected from \{10, 20, 30, 40, 50, 60, 70, 80, 90, 100\}, \{0.01, 0.1, 0.5, 1, 2, 5, 10, 20, 50\} and \{0.01, 0.05, 0.1, 0.5, 1, 2, 5, 10\}, respectively.

\subsection{Experiment Results}
Table~\ref{tab2} exhibits the clustering results of baselines and JPLTD on seven popular datasets with various missing ratios. From the experimental results, it can be observed that:

\textit{1)} JPLTD outperforms the baselines on all datasets with varying missing rates. Especially, JPLTD leaps forward on RGB-D and Scene datasets, compared with the second-best approach. Taking the results on RGB-D and Scene datasets with 50\% missing rate for example. On RGB-D dataset, JPLTD achieves 16.60\%, 25.29\%, and 20.89\% average improvements over the second-best method (i.e., IMVC-CBG) in terms of ACC, NMI, and ARI, respectively. On Scene dataset, our method obtains 27.29\%, 32.77\%, and 41.43\% average improvements compared with the second-best method (i.e., IMVTSC-MVI) in terms of ACC, NMI, and ARI, respectively. These results validate the effectiveness of JPLTD for incomplete multi-view data clustering.

\textit{2)} Generally, since the loss of view information has a negative impact on the excavation of complementary information across views,  the clustering performance of all methods will become worse with the increase of the missing rate. However, when the missing rate increases from 0.1 to 0.7, the performance drop of JPLTD is small. Taking ACC for instance, when the missing rate increases by 0.2, the average decrease of experimental performance on seven datasets (i.e., ORL, NGs, MSRC, BBCSport, RGB-D, UCI, Scene) are 0.66\%, 0.13\%, 0.24\%, 0.34\%, 1.01\%, 0.10\%, and 0.32\%, respectively. This demonstrates the robustness of  JPLTD against the missing views.

\textit{3)} There are five graph-based methods used for comparison, i.e., IMVTSC-MVI, TMBSD, TC-IMVC, VCL-IMVC, and IMVC-CBG. Compared with these methods, JPLTD not only considers the redundant features and noise in the original high-dimensional data for projection learning but also utilizes tensor learning to take the noise introduced by graph learning into account. The experimental results verify that the consideration of projection learning and the reduction of graph noise is important for the exploration of latent information and the intrinsic similarity of incomplete multi-view data.

\begin{figure*}[t]
	\centering
	\begin{minipage}[c]{1\textwidth}
		\centering
		\includegraphics[width=18cm,height=3.4cm]{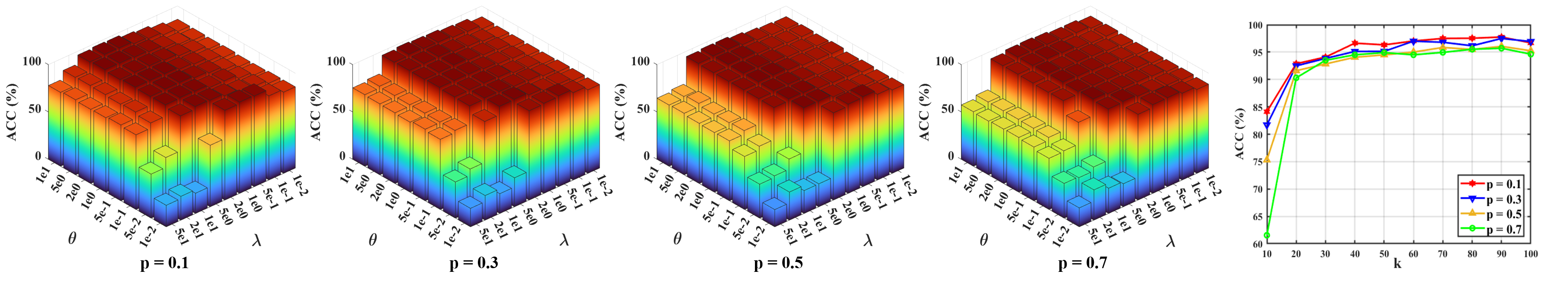}
		\centerline{(a)\ On ORL}
	\end{minipage}
	\begin{minipage}[c]{1\textwidth}
		\centering
		\includegraphics[width=18cm,height=3.4cm]{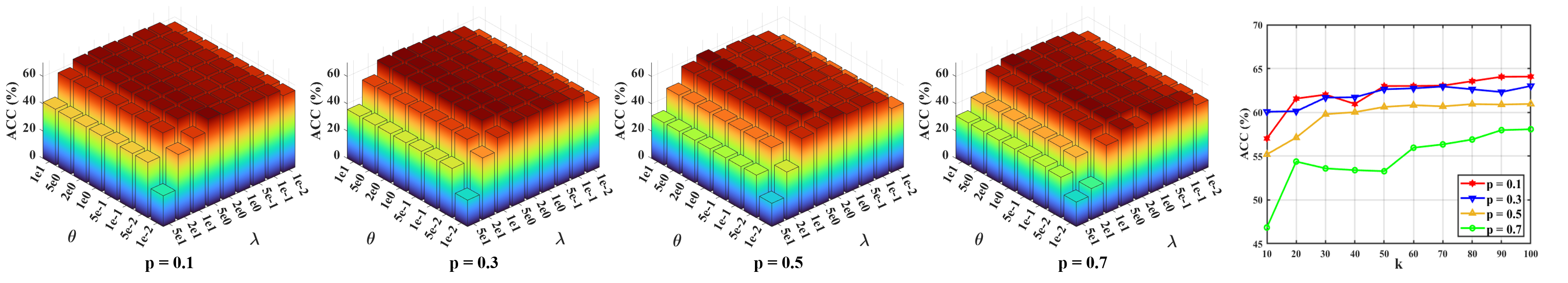}
		\centerline{(b)\ On RGB-D}
	\end{minipage}
	\caption{ACC values of JPLTD w.r.t $\lambda$, $\theta$, and $k$  on ORL dataset and RGB-D dataset with different missing rate \textit{p}. }
	\label{fig4}
\end{figure*}
\begin{figure*}[t]
	\centering
	\begin{minipage}[c]{0.245\textwidth}
		\centering
		\includegraphics[width=4.5cm,height=3.9cm]{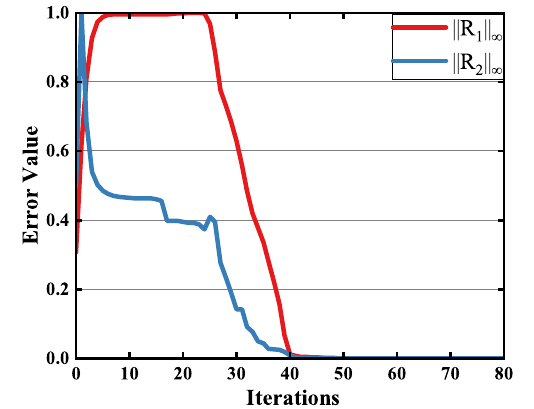}
		\centerline{(a)\ ORL}
	\end{minipage}
	\begin{minipage}[c]{0.245\textwidth}
		\centering
		\includegraphics[width=4.5cm,height=3.9cm]{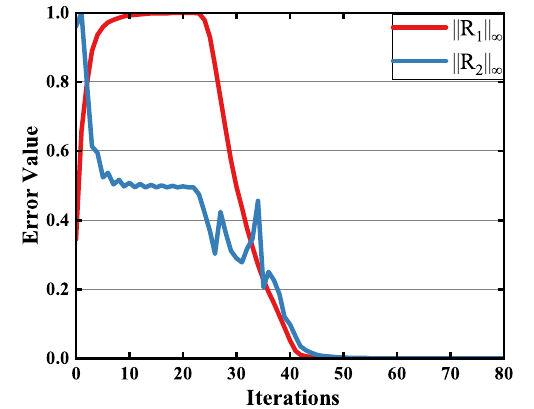}
		\centerline{(b)\ NGs}
	\end{minipage}
	\begin{minipage}[c]{0.245\textwidth}
		\centering
		\includegraphics[width=4.5cm,height=3.9cm]{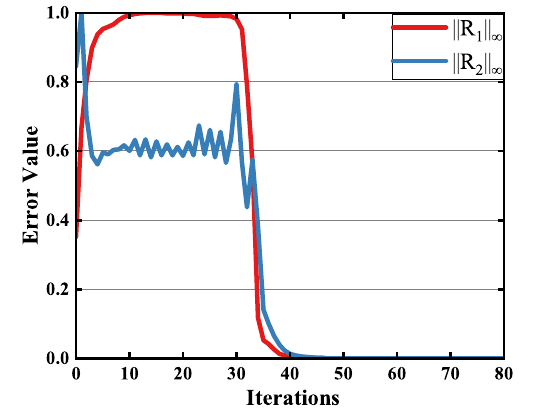}
		\centerline{(c)\ BBCSport}
	\end{minipage}
	\begin{minipage}[c]{0.245\textwidth}
		\centering
		\includegraphics[width=4.5cm,height=3.9cm]{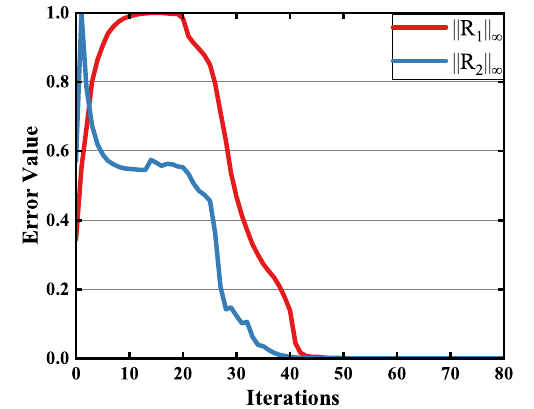}
		\centerline{(d)\ RGB-D}
	\end{minipage}\\[3mm]
	\caption{Convergence curves of JPLTD on ORL, NGs, BBCSport, and RGB-D with  missing rate $p$ = 0.7. }
	\label{fig5}
\end{figure*}

\subsection{Visualization}
JPLTD learns an intrinsic tensor $\mathcal{B}$ and a sparse tensor $\mathcal{P}$ from the origin tensor $\mathcal{G}$ for clustering. To intuitively show the effectiveness of graph recovery of JPLTD, we provide a visualization of the learned graph structure of JPLTD. Fig.~\ref{fig3} exhibits the learned similarity graphs of JPLTD on ORL and Scene datasets with three types of missing rates, where the number of diagonal blocks is equal to the cluster number. For the sake of clarity, we select similar graphs of ten classes of samples on ORL dataset for visualization. Besides, we also display the similarity graphs of a competitive method (i.e., IMVTSC-MVI) in Fig.~\ref{fig3} for comparison. We can see that JPLTD reveals the block diagonal structure of graphs well even if the missing rate is equal to 0.7. Comparatively, the structure of similarity graphs learned by IMVTSC-MVI is sparse especially when facing the Scene dataset, indicating that it cannot excavate the relevance of samples within a class well. Compared with IMVTSC-MVI, the graphs learned by JPLTD have stronger inter-class discrepancy and intra-class compactness in different missing rates. These visualization results verify that our method can discover the intrinsic similarity among incomplete multi-view samples effectively.

\subsection{Parameter Analysis}
The proposed JPLTD model has three tradeoff parameters that  affect the clustering performance, i.e., $\lambda$, $\theta$, and the dimension of projected space $k$. Herein, we investigate the parameter sensitivity of JPLTD w.r.t these parameters. We first define a candidate
set \{1e-2, 1e-1, 5e-1, 1e0, 2e0, 5e0, 1e1, 2e1, 5e1\} for $\lambda$, \{1e-2, 5e-2, 1e-1, 5e-1, 1e0, 2e0, 5e0, 1e1\} for $\theta$, and \{10, 20, ..., 100\} with stepsize 10 for $k$.
Then, with various combinations of parameters $\lambda$, $\theta$, and $k$, we conduct experiments on ORL and RGB-D datasets with four missing rates. Fig.~\ref{fig4} shows the
change of ACC value of JPLTD versus $\lambda$, $\theta$, and $k$.  It can be observed that our method is more sensitive to $\lambda$ than $\theta$. Since $\lambda$ controls the weight of the intrinsic tensor, whose quality is closely related to the final clustering effect, the learned intrinsic tensor may be biased if $\lambda$ is inappropriate, leading to a reduction in clustering performance. Generally, JPLTD can obtain satisfactory results when $\lambda$ is selected from [2e0, 2e1], $k$ is selected from [60, 100], and $\theta$ is selected from [5e-2, 2e0] for all datasets.
\begin{table*}[t]
	\scriptsize
	\centering
	\caption{Ablation Study Results on Five different datasets.}
	\label{tab3}
	\renewcommand{\arraystretch}{1.1}
	\scalebox{0.97}{ 
	\begin{tabular}{|p{0.1cm}|p{1.5cm}|p{0.87cm} p{0.87cm} p{1.0cm} | p{0.87cm} p{0.87cm} p{1.0cm} |p{0.87cm}p{0.87cm} p{1.0cm}|p{0.87cm}p{0.87cm} p{1.0cm}|}
		\hline
		& Missing rates&\multicolumn{3}{c}{0.1}\vline & \multicolumn{3}{c}{0.3}\vline & \multicolumn{3}{c}{0.5}\vline& \multicolumn{3}{c}{0.7}\vline \\
		\hline
		& Metrics  & ACC~(\%)   & NMI~(\%)   & ARI~(\%)   &     ACC~(\%)   & NMI~(\%)   & ARI~(\%) &    ACC~(\%)   & NMI~(\%)   & ARI~(\%)   &  ACC~(\%)   & NMI~(\%)   & ARI~(\%) \\
		\hline
		\multirow{4}{*}{\rotatebox{90}{\textbf{ORL}}}
		&JPLTD-N	&96.76±1.31	&99.34±0.27	&96.78±1.29	&96.26±1.30	&99.04±0.29	&96.00±1.27	&94.90±2.10	&98.38±0.46	&93.99±1.85	&93.88±1.79	&97.88±0.64	&92.39±2.16\\
		&JPLTD-B	&95.91±1.40	&98.89±0.41	&95.46±1.40	&95.28±2.34	&98.79±0.57	&94.99±2.33	&92.29±1.98	&97.59±0.59	&91.06±2.03	&87.81±2.74	&94.66±1.21	&84.02±3.26\\ 
		&JPLTD-O    &93.65±1.95	&98.57±0.46	&93.50±1.83	&92.64±2.35	&98.00±0.56	&91.91±2.29	&90.15±2.45	&96.84±0.58	&88.50±2.24	&87.34±2.76	&94.40±1.20	&83.51±3.32\\	     
		&\textbf{JPLTD}	&\textbf{97.65±1.72}&\textbf{99.47±0.38}&\textbf{97.50±1.73}&\textbf{97.51±1.50} &\textbf{99.43±0.32}&\textbf{97.46±1.51}&\textbf{96.36±1.33}&\textbf{98.68±0.32}&\textbf{95.29±1.36}&\textbf{95.66±1.72}&\textbf{98.62±0.45}&\textbf{95.03±1.76}\\
		\hline
		\multirow{4}{*}{\rotatebox{90}{\textbf{NGs}}}
		&JPLTD-N	&99.60±0.00	&98.61±0.00	&99.00±0.00	&99.60±0.00	&98.61±0.00	&99.00±0.00	&99.60±0.00	&98.61±0.00	&99.00±0.00	&99.20±0.00	&97.39±0.00	&98.00±0.00\\
		&JPLTD-B	&98.60±0.00	&95.30±0.00	&96.52±0.00	&98.60±0.00	&95.30±0.00	&96.52±0.00	&98.00±0.00	&93.62±0.00	&95.02±0.00	&95.00±0.00	&85.75±0.00	&87.88±0.00\\
		&JPLTD-O	&98.00±0.00 &93.80±0.00	&95.06±0.00	&98.00±0.00	&94.38±0.00	&95.04±0.00	&97.20±0.00	&91.26±0.00	&93.10±0.00	&93.80±0.00	&81.98±0.00	&85.14±0.00\\		
		&\textbf{JPLTD}	&\textbf{100.0±0.00}	&\textbf{100.0±0.00}	&\textbf{100.0±0.00}	&\textbf{99.80±0.00}	&\textbf{99.30±0.00}	&\textbf{99.50±0.00}	&\textbf{99.60±0.00}	&\textbf{98.61±0.00}	&\textbf{98.99±0.00}	&\textbf{99.60±0.00}	&\textbf{98.61±0.00}	&\textbf{98.99±0.00}\\
		\hline		
			\multirow{4}{*}{\rotatebox{90}{\textbf{MSRC-v1}}}
		&JPLTD-N	&95.12±0.21	&89.77±0.34	&88.80±0.51	&93.67±0.22	&87.76±0.51	&85.94±0.55	&92.38±0.29	&84.31±0.18	&82.58±0.29	&87.14±0.32	&79.89±0.30	&74.81±0.64\\
		&JPLTD-B	&93.83±0.11	&88.56±0.13	&86.07±0.22	&91.88±0.11	&85.83±0.19	&82.46±0.20	&87.62±0.48	&83.72±0.61	&77.81±0.83	&83.76±0.17	&75.66±0.39	&68.77±0.40\\
		&JPLTD-O	&86.19±0.29	&79.64±0.24	&74.63±0.14	&82.69±0.32	&75.10±0.51	&67.27±0.62	&76.45±0.89	&71.27±0.46	&63.76±0.23	&70.76±0.47	&64.09±0.56	&52.96±0.59\\		
		&\textbf{JPLTD}	&\textbf{100.0±0.00} &\textbf{100.0±0.00}	&\textbf{100.0±0.00}	&\textbf{99.52±0.00}	&\textbf{98.92±0.00}	&\textbf{98.88±0.00}	&\textbf{99.28±0.24}	&\textbf{98.38±0.54}	&\textbf{98.31±0.57}	&\textbf{99.27±0.24}	&\textbf{98.36±0.54}	&\textbf{98.30±0.56}\\
		\hline
		\multirow{4}{*}{\rotatebox{90}{\textbf{BBCSport}}}
		&JPLTD-N	&98.90±0.00	&96.28±0.00	&96.86±0.00	&98.90±0.00	&96.20±0.00	&97.07±0.00	&98.34±0.00	&94.52±0.00	&95.60±0.00	&97.24±0.00	&91.71±0.00	&93.63±0.00\\
		&JPLTD-B	&97.98±0.00	&93.28±0.00	&94.65±0.00	&97.61±0.00	&92.93±0.00	&95.23±0.00	&96.14±0.00	&93.12±0.00	&95.23±0.00	&95.22±0.00	&90.26±0.00	&92.35±0.00\\
		&JPLTD-O   &96.51±0.00	&92.70±0.00	&94.80±0.00	&96.14±0.00	&87.63±0.00	&89.62±0.00	&95.58±0.00	&87.59±0.00	&89.84±0.00	&93.58±0.00	&88.07±0.00	&89.31±0.00\\
		&\textbf{JPLTD}	&\textbf{99.91±0.09} &\textbf{99.67±0.34}	&\textbf{99.83±0.17}	&\textbf{99.63±0.19}	&\textbf{98.69±0.70}	&\textbf{99.32±0.45}	&\textbf{99.45±0.19}	&\textbf{98.13±0.57}	&\textbf{98.76±0.55}	&\textbf{98.90±0.37}	&\textbf{96.50±0.92}	&\textbf{97.30±0.96}\\	
		\hline
		\multirow{4}{*}{\rotatebox{90}{\textbf{RGB-D}}}
		&JPLTD-N	&63.19±0.12	&60.54±0.17	&47.35±0.15	&60.98±0.14	&58.65±0.13	&44.96±0.15	&58.44±0.23	&52.42±0.16	&40.08±0.25 &54.45±0.73	&51.52±0.25	&36.70±0.66\\
		&JPLTD-B	&61.02±0.43	&60.09±0.35	&45.38±0.36	&57.95±0.13	&54.66±0.14	&41.00±0.19	&53.25±0.59	&50.44±0.31	&36.10±0.30	&50.22±0.26	&41.83±0.37	&32.19±0.39\\	
		&JPLTD-O    &58.57±0.12	&60.58±0.16	&44.98±0.16	&56.39±0.40 &53.13±0.26	&40.67±0.17	&52.38±0.62	&49.80±0.43	&36.95±0.27	&49.41±0.32	&41.83±0.32	&31.97±0.40\\		
		&\textbf{JPLTD}	&\textbf{64.73±0.04}	&\textbf{62.26±0.10}	&\textbf{50.10±0.10}	&\textbf{62.52±0.08}	&\textbf{59.65±0.13}	&\textbf{46.16±0.12}	&\textbf{62.29±0.06}	&\textbf{59.25±0.08}	&\textbf{45.45±0.08}	&\textbf{61.70±0.33}	&\textbf{55.23±0.18}	&\textbf{43.72±0.19}\\		
		\hline
	\end{tabular}%
}
	\label{Core}%
\end{table*}%
 \begin{figure}[!t]
	\centering
	\includegraphics[width=8.67cm,height=3.25cm]{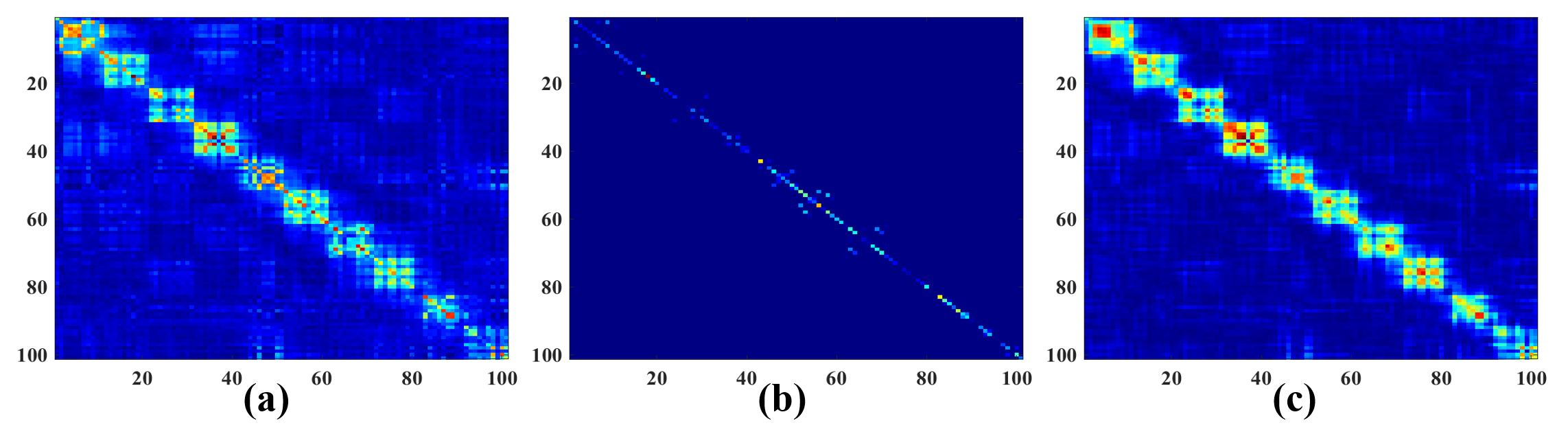}
	\caption{An example to explain the recovered graphs of JPLTD-B and JPLTD on ORL dataset with missing rate $p$ = 0.5. Figure~(a) is the recovered graph of JPLTD-B. Figure~(b) is the learned sparse noise graph of JPLTD, and (c) is the learned intrinsic graph of JPLTD.}
	\label{fig6}
\end{figure}
\subsection{Convergence Analysis}
In this subsection, we experimentally investigate the convergence property of our optimization algorithm. Specifically, we adopt the primal residuals as convergence conditions and let $R_{1}$and $R_{2}$ denote $(\textbf{W}_{t+1}^{v}\textbf{X}_{o}^{v} - \textbf{W}_{t+1}^{v}\textbf{X}_{o}^{v}\textbf{A}^{v}\textbf{G}_{t+1}^{v}\textbf{A}^{v^{T}}-\textbf{E}_{t+1}^{v})$ and $(\mathcal{G}_{t+1}-\mathcal{B}_{t+1}-\mathcal{P}_{t+1})$, respectively. Fig.~\ref{fig5} shows the residual values in each iteration on ORL, NGs, BBCSport, and RGB-D datasets with missing rate $p$ = 0.7, where the error value under each experiment is normalized by dividing the maximum value. We can observe that all the residuals can converge to 0 in less than 80 iterations, indicating that our algorithm  has good convergence properties.

\subsection{Ablation Study}
JPLTD projects the original data into a lower-dimensional space for graph learning, and it decomposes the initial tensor into an intrinsic tensor and a sparse tensor to better depict the similarities of incomplete data. Herein, we conduct an ablation study to investigate the effectiveness of projection learning and tensor decomposition in JPLTD. We derive three variants  based on the original JPLTD model, i.e., JPLTD-N, JPLTD-B, and JPLTD-O. JPLTD-N  gives up the dimension reduction of the original data, its loss function is:
\begin{equation}\label{eq24}
\begin{aligned}
&\min\limits_{\Lambda}\ \sum_{v=1}^{m}\|\textbf{E}^{v}\|_{2,1} + \lambda \|\mathcal{B}\|_{\circledast}+ \theta\|\mathcal{P}\|_{1}\\
&s.t.\ \textbf{E}^{v} = \textbf{X}_{o}^{v} - \textbf{X}_{o}^{v}\textbf{A}^{v}\textbf{G}^{v}\textbf{A}^{v^{T}},\ \mathcal{G} = \mathcal{B}+\mathcal{P},\ \\
 &\mathcal{G} = \Psi(\textbf{G}^{1},\ \textbf{G}^{2},\cdots,\ \textbf{G}^{m}),
\end{aligned}
\end{equation}
JPLTD-B drops the  sparse tensor term and directly utilizes the original tensor $\mathcal{G}$ for clustering, its loss function can be described as:
\begin{equation}\label{eq25}
\begin{aligned}
&\min\limits_{\Lambda}\ \sum_{v=1}^{m}\|\textbf{E}^{v}\|_{2,1} + \lambda \|\mathcal{G}\|_{\circledast}\\
s.t.\ &\textbf{E}^{v} = \textbf{W}^{v}\textbf{X}_{o}^{v} - \textbf{W}^{v}\textbf{X}_{o}^{v}\textbf{A}^{v}\textbf{G}^{v}\textbf{A}^{v^{T}},\ \textbf{W}^{v}\textbf{W}^{v^{T}} = \textbf{I}_{k},\\ 
&\mathcal{G} = \Psi(\textbf{G}^{1},\ \textbf{G}^{2},\cdots,\ \textbf{G}^{m}),
\end{aligned}
\end{equation}
JPLTD-O considers neither the dimension reduction of the original data nor the sparse tensor term, its loss function is:
\begin{equation}\label{eq26}
\begin{aligned}
&\min\limits_{\Lambda}\ \sum_{v=1}^{m}\|\textbf{E}^{v}\|_{2,1} + \lambda \|\mathcal{G}\|_{\circledast}\\
s.t.\ \textbf{E}^{v} = &\textbf{X}_{o}^{v} -\textbf{X}_{o}^{v}\textbf{A}^{v}\textbf{G}^{v}\textbf{A}^{v^{T}}, \ 
\mathcal{G} = \Psi(\textbf{G}^{1},\ \textbf{G}^{2},\cdots,\ \textbf{G}^{m}),
\end{aligned}
\end{equation}

Table~\ref{tab3} exhibits the ACC, NMI, and ARI results of JPLTD and its variants on four datasets (i.e., ORL, NGs, BBCSport, RGB-D) with different missing rates. We can observe that both JPLTD-N and JPLTD-B are superior to JPLTD-O and JPLTD improves the performance over JPLTD-N, JPLTD-B, and JPLTD-O on all datasets, which indicates the noise in the original high-dimensional data or the original tensor  would degrade the clustering performance. 

To further investigate the effectiveness of the denoising module, we exhibit the recovered graphs of JPLTD-B and JPLTD on ORL dataset in Fig.~\ref{fig6}. Fig.~\ref{fig6}(a) is the recovered graph of JPLTD-B. Fig.~\ref{fig6}(b) is the learned sparse noise graph of JPLTD, and (c) is the learned intrinsic graph of JPLTD. It can be observed that the block diagonal structure shown by (c) is more clear than that of (a) and the sparse noise is alleviated, which indicates the denoising module in our model is beneficial for exploring the  intrinsic similarity among data.
These experimental results verify the effectiveness of projection learning and tensor decomposition.

\section{Conclusion}
In this paper, we propose a novel Joint Projection learning and Tensor Decomposition (JPLTD) based method for incomplete multi-view clustering, which integrates graph learning and tensor recovery into a unified model.  On the one hand, JPLTD projects the initial high-dimensional data into a lower-dimensional space for compact feature learning to alleviate the influence of original feature redundancy. On the other hand, to reduce the influence of sparse noise introduced by graph learning, JPLTD decomposes the original tensor formed by incomplete graphs into an intrinsic tensor and a sparse tensor. The sparse tensor describes the sparse noise. The intrinsic tensor models the true data similarities and is utilized for final similarity graph construction.
An effective algorithm is introduced for solving the JPLTD model.  Extensive experimental results on several popular datasets validate that JPLTD outperforms the state-of-the-art IMVC approaches.
A limitation of our method is that JPLTD will encounter high complexity and time consumption when applied to large-scale datasets. In the future, we will consider using the anchor graph to reduce its complexity while maintaining its good performance.

\ifCLASSOPTIONcaptionsoff
  \newpage
\fi



%

\bibliographystyle{IEEEtran}
\bibliography{IEEEabrv,bibfile}
%
%

%

%
%
%




\end{document}